\documentclass{article}

\usepackage[english]{babel}
\usepackage{graphicx}
\usepackage{float}
\usepackage{amsmath}
\usepackage{amssymb}
\usepackage{amsfonts}
\usepackage{soul}
\usepackage{xcolor}

\usepackage{booktabs}
\usepackage[hidelinks,colorlinks=true,linkcolor=blue,citecolor=blue]{hyperref}
\usepackage[all]{hypcap}
\usepackage{todonotes}
\usepackage{tabularx}
\usepackage{subcaption}
\usepackage{appendix}
\usepackage[mathscr]{euscript}
\usepackage{lineno}
\usepackage{csquotes}
\usepackage[backend=biber,style=numeric-comp,sorting=none]{biblatex}
\bibliography{bibliography.bib}
\DeclareFieldFormat[article]{title}{#1}
\renewbibmacro{in:}{}
\usepackage{authblk}

\usepackage[a4paper, left=1in, right=1in, top=1in, bottom=1in]{geometry}

\title{{Kolmogorov Arnold networks (KAN) for aerodynamic prediction: a comparison with MLPs and GNNs}}

\author[1]{Miguel Jaraiz}
\author[1]{Fermin Gutierrez}
\author[1]{Pablo Yeste}
\author[1]{Miguel Sánchez-Domínguez}
\author[1,3]{Eusebio Valero}
\author[1,3]{Gonzalo Rubio}
\author[2]{Lucas Lacasa}

\affil[1]{ETSIAE-UPM-School of Aeronautics, Universidad Politécnica de Madrid, Plaza Cardenal Cisneros 3, E-28040 Madrid, Spain}
\affil[2]{Institute for Cross-Disciplinary Physics and Complex Systems (IFISC, CSIC-UIB), 07122 Palma de Mallorca (Spain)}
\affil[3]{Center for Computational Simulation, Universidad Politécnica de Madrid, Campus de Montegancedo, Boadilla del Monte, 28660 Madrid, Spain}

\date{\today}

\begin{document}

\maketitle
%\linenumbers

%\newpage

\begin{abstract}
{Kolmogorov Arnold networks (KAN) have recently been introduced as a (deep) neural network architecture whose trainable parameters adapt the activation functions, instead of the coefficients of the affine transformations at the core of traditional architectures such as deep multilayer perceptrons (MLPs). This architecture builds on the Kolmogorov-Arnold theorem, which endows it with universal approximation properties. While the advent of KANs has been received with excitement, there is a current debate about the possible KAN supremacy over deep multilayer perceptrons (MLPs) for classic fields such as symbolic regression, generic-purpose machine learning, natural language processing or computer vision. Here we assess the performance of KANs --and its nuanced comparison against MLPs and graph neural networks (GNNs)-- in the realm of fluid dynamics surrogate modelling. To that aim, we consider the task of predicting the surface pressure distribution over subsonic and transonic airfoils, a canonical task in aerodynamics. Our results show that KAN models show good performance in predicting the whole pressure coefficients and is able to interpolate across Mach numbers and angles of attack, however its performance is comparable --marginally inferior-- to a suitably trained MLP, where best performance is achieved by a GNN at the expense or requiring lengthier training. While the optimal KAN model have typically much lower complexity than MLP and GNN --hence resulting in faster training--, we find that KANs suffer from training instabilities, and their performance is highly dependent on a proper hyperparameter optimisation.}
\end{abstract}

\section{Introduction}
The simulation of physical processes across a wide range of disciplines boils down to numerically solving partial differential equations (PDE) over a specific domain geometry and a range of physical parameters. A paradigmatic case is Computational Fluid Dynamics (CFD) \cite{blazek2015computational} --a fundamental simulation tool e.g. for aircraft design-- which proceeds to integrate flow equations of motion (i.e. Navier-Stokes equations or adequate approximations such as Reynolds-Averaged Navier-Stokes (RANS) or Large Eddy Simulations (LES)). CFD is often seen as the gold standard for the numerical analysis of fluid-dynamical problems as it provides high-fidelity solutions of complex flow phenomena. However, CFD-based methods
%which can be used for accurate aerodynamic analysis. Techniques like Reynolds-Averaged Navier-Stokes (RANS) and Large Eddy Simulations (LES) are commonly employed to capture complex flow phenomena, but these methods 
are computationally expensive, both in terms of simulation time and required computational resources. For example, when estimating aerodynamic loads, each unique flight condition within an aircraft's operational envelope requires a separate CFD simulation, leading to an immense computational burden.

\medskip \noindent
Surrogate models \cite{brunton2022data} provide an efficient alternative, allowing for rapid approximation of CFD results across varying conditions without the need for costly repeated simulations.
Traditional surrogate modeling has been approached through data-driven methods such as proper orthogonal decomposition (POD) combined with interpolation techniques \cite{iuliano2013proper,franz2014interpolation,fossati2015evaluation} or dynamic mode decomposition (DMD) \cite{brunton2022data}. These models are relatively straightforward to construct and perform well for predictable linear behavior, but struggle to accurately approximate intrinsically nonlinear behavior. The advent of powerful machine learning methods, particularly deep supervided learning (DL) \cite{Goodfellow-et-al-2016} has opened in recent years a new approach to construct surrogate models \cite{brunton2022data}.
%However, POD struggles with flow field predictions at transonic conditions, where nonlinear phenomena make it difficult to approximate flow fields as linear superpositions of a limited set of modes. This limitation has driven research into alternative modeling approaches.
%In recent years, deep learning (DL) models have garnered significant attention in this area. 
In the context of fluid-dynamical surrogate models, several studies have demonstrated the potential of machine learning for reduced-order modeling \cite{brunton2020machine, le2023improving, kou2017layered, jin2018prediction, bhatnagar2019prediction, wu2020deep, thuerey2020deep, rozov2021data, soler2024reinforcement, ramos2025transfer, kou2023aeroacoustic, ladron2025certifiable}, including not only academically-oriented models, but also realistic, industrial-level 3D geometries, where unstructured grids with millions of points are necessary to accurately capture aircraft configurations.\\
A paradigmatic problem in aerodynamic CFD with crucial industrial interest is the prediction of aerodynamic coefficients over the surface of an aircraft, e.g. the prediction of pressure coefficients over an airfoil for specific physical parameters (Reynolds or Mach number) and flight conditions (e.g. angle of attack). Comparative studies highlight various techniques in surrogate regression models for aerodynamic coefficient prediction \cite{andres2021application}. For instance, in \cite{sabater2022fast} a coordinate-based MLP --the default DL architecture for supervised learning regression tasks-- was proposed for the pointwise prediction of such pressure coefficients. Graph neural networks (GNN) have also been proposed as more sophisticated supervised learning architectures for this task  \cite{hines2023graph}, and more recently generative AI (flow matching) has also been repurposed for this task \cite{ramos2026fluidflow}. 
%Finally, other learning paradigms such as reinforcement or transfer learning have also been applied to geometric optimisation of airfoils \cite{ramos2025transfer}. %Recently, \cite{catalani2024neural} presented a methodology based on Implicit Neural Representations (INRs) to learn surrogate models of steady-state fluid dynamics simulations on meshed domains, demonstrating some improvements over GNN-based models.

\medskip \noindent
Within supervised learning, observe that both MLPs and GNNs rely on universal approximation theorems \cite{hornik1991approximation} to approximate nonlinear functions in terms of an sequential composition of affine combinations and nonlinear activation functions. One of the latest advancements in this field is the inception of so-called Kolmogorov–Arnold Networks (KANs) \cite{liu2024kan}. At odds with previous architectures, KANs rely on a different representation theorem  (Kolmogorov-Arnold representation theorem \cite{kolmogorov1961representation, ismailov2025addressing}) and were introduced as an alternative to MLPs where, unlike MLPs, KANs feature learnable activation functions on edges, promising greater flexibility and interpretability. 
%This seemingly simple adjustment enables KANs to outperform MLPs in terms of both accuracy and interpretability.
Since their emergence, KANs have received wide popularity and shown great promise in a number of applications, both in science and industry \cite{somvanshi2025survey}. The supremacy of KANs over MLPs is, however, currently a matter of debate \cite{shukla2024comprehensive}. For instance, recent works \cite{yu2024kan} argue that MLP might actually remain superior than KANs over a large set of tasks including machine learning, computer vision, audio processing, or natural language processing. In this context, it is unclear whether KANs generally shows higher performance --as compared to MLPs or related architectures-- for the design of surrogate models of physical processes \cite{kashefi2024kolmogorov}.
%On the one hand, the derivation of activation functions and their combination with each other provides a numerical aid to obtain an analytical solution of unsolved mathematical problems. On the other hand, the linear combination of non-linear functions provides greater adaptability of the network to non-linear problems, unlike traditional MLPs, which combine linear functions for the same type of problems. 
%This is the case of the Navier-Stokes equations, where there are already studies of the application of these networks in flow around complicated geometries and obtaining CFD solutions \cite{kashefi2024kolmogorov}.

\medskip \noindent
In this paper, we build a KAN-based surrogate model for a vanilla aerodynamic task: the prediction of pressure coefficients over a boundary airfoil, and compare its performance in interpolating across physical parameters and flight conditions (Mach number, angle of attack) against two well-established techniques: MLPs \cite{sabater2022fast} and GNNs \cite{hines2023graph}. We build on \cite{hines2023graph}, replicate their results on the performance of both MLPs and GNNs on the same learning task and then compare these against the KAN-based surrogate model.\\
%Note that the prediction of aerodynamic magnitudes over a RANS mesh has been previously studied as a curve-fitting problem \cite{hines2023graph}. We follow established conventions and adopt the problem setting and dataset from \cite{hines2023graph} and assess the performance of a KAN-based surrogate model in comparison to MLP and GNN-based surrogates. 
Our results indicate that while all three models achieve good results (where GNN reaches the best performance), the KAN-based model ranks third, achieving comparable yet marginally inferior performance with respect to its MLP-based counterpart, and suffers from training instabilities and poorer generalisation error in the more difficult regions where the pressure coefficients show stronger gradients. At the same time, training of KAN model converges substantially faster --i.e. needing fewer training epochs-- than the MLP and GNN, arguably because the complexity of optimal MLP and GNN architectures is substantially larger (in terms of number of trainable parameters) than the optimal KAN architecture. This effect has recently been observed in a KAN vs MLP comparison within reinforcement learning \cite{kich2024kolmogorov}.
Now, by training smaller MLPs --with a complexity comparable or even smaller than the KAN model--, we show that we can still reach generalisation performance on par with KANs. Despite the recent popularity of KANs, these results overall challenge their acclaimed supremacy over more traditional deep learning architectures.\\
The rest of the paper is organised as follows. In Section \ref{sec:methodology}, we introduce the entire methodology: the learning task, the data, and details on all three surrogate models. Section \ref{sec:results} reports our results, including a comprehensive hyperparameter optimisation analysis --highlighting the strong dependency of KAN's performance on a proper choice of hyperparameters, even on the random seed \cite{picard2021torch}--, the comparison of the performance for optimal MLPs, GNNs and KANs in terms of learning curve convergence, pointwise test error metrics and preservation of correlation structures. In this section we also discuss how MLP and KAN of comparable complexities compare in terms of performance. Finally, in Section \ref{sec:conclusions} we conclude.

\section{Methodology} \label{sec:methodology}

%\subsection{Problem under study in a nutshell} \label{subsec:problem_under}
\subsection{Problem setup and data preparation}
The learning task considered here is identical across all data‐driven models: given a vector of operating conditions, predict the distribution of pressure coefficients over an airfoil via a RANS-based CFD simulation. To that aim, the airfoil geometry is first placed inside a computational domain and discretised with an unstructured mesh that resolves the boundary layer near the surface and extends to a far-field boundary representing the freestream conditions. The RANS equations (together with a turbulence model) are then discretised on this mesh using a finite-volume method and solved iteratively until the velocity and pressure fields converge. Once the solution is obtained, the static pressure values at the nodes/faces that lie on the airfoil boundary are extracted from the computed flow field. These pressures are then converted into pressure coefficients by subtracting the freestream pressure and normalizing by the freestream dynamic pressure. Mapping these values along the airfoil surface as a function of the chordwise coordinate produces the pressure-coefficient distribution $C_p(x)$ on the airfoil.

\begin{figure}[htb!]
    \centering
    \includegraphics[width=0.6\linewidth]{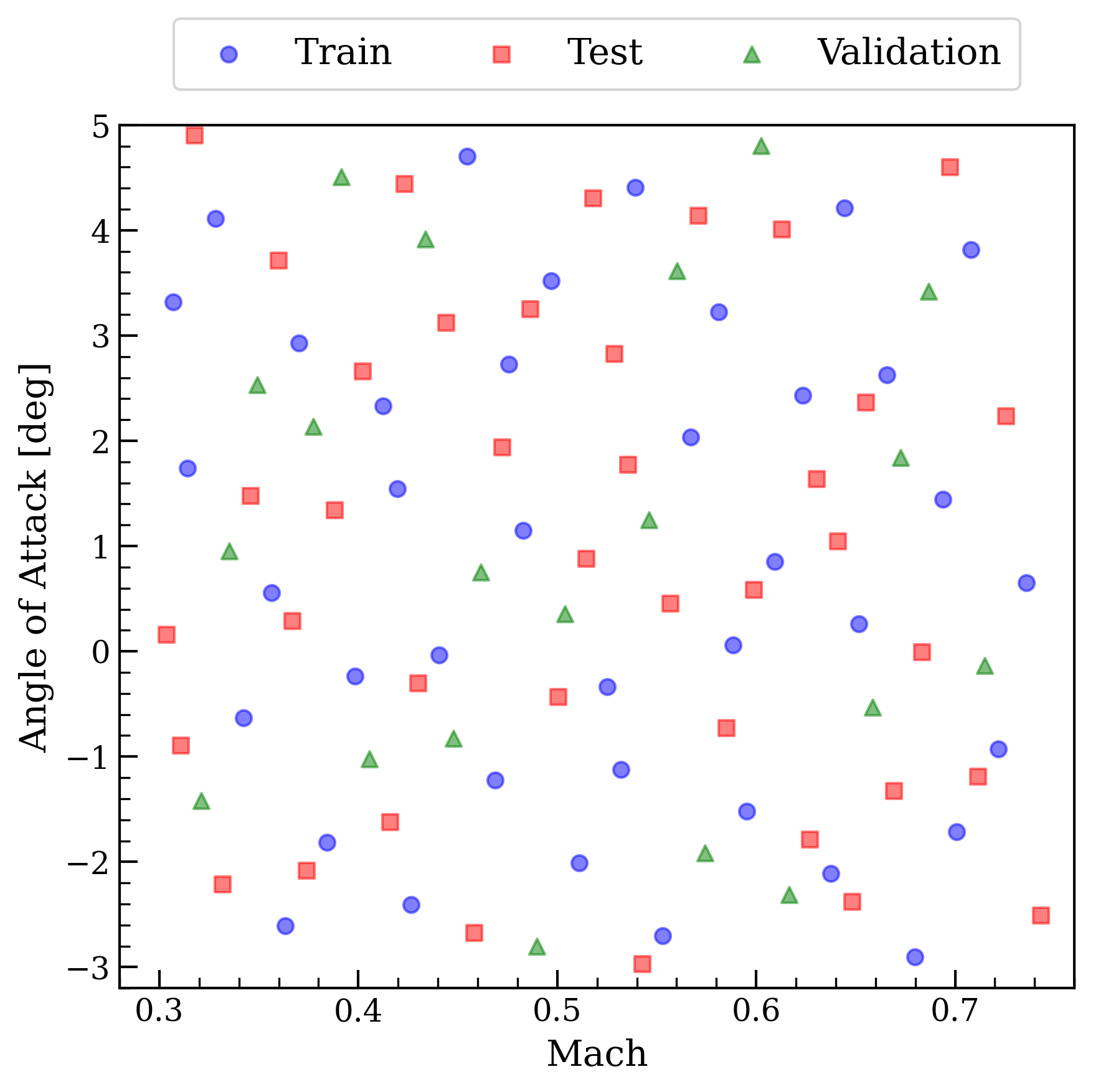}
    \caption{Data split.}
    %\begin{subfigure}{0.525\linewidth}
    %    \centering
    %    \includegraphics[width=\linewidth]{fig/NLR7301mesh.png}
    %    \caption{Mesh}
    %    \label{fig:NLR7301mesh}
    %\end{subfigure}
    %\hfill
    %\begin{subfigure}{0.465\linewidth}
    %    \centering
    %    \includegraphics[width=\linewidth]{fig/dataset_split.png}
    %    \caption{Data split.}
    %    \label{fig:datasplit_NGval_DLR}
    %\end{subfigure}
    %\caption{NLR7301 dataset, as provided by \textcite{hines2023graph}. \textcolor{red}{ESTAS FIGURAS SON NUESTRAS O COPY-PASTEADAS? HABRÍA QUE HACER LAS NUESTRAS} \textcolor{magenta}{la malla no la tenemos, pero podemos pintar el perfil o el perfil junto con una distribución de presión de ejemplo. El splitting es facil generarlo.} \textcolor{orange}{FG: la del split es nuestra, de del perfil no}}
    \label{fig:nlr7301_dataset}
\end{figure}

\medskip\noindent The database available for this study was kindly shared by DLR \cite{hines2023graph}, but we nonetheless provide details on how this was constructed for the sake of self-containedness.\\
As the surface geometry, we use the supercritical NLR7301 airfoil introduced in \textcite{hines2023graph}. 
High-fidelity RANS-based CFD computations were performed with the DLR flow solver TAU \cite{schwamborn2006dlr}, employing the Spalart-Allmaras turbulence model.
The computational domain is discretised with an unstructured mesh of $29,441$ nodes,  where a total of $n=597$ nodes are located on the airfoil boundary. 
%(see \autoref{fig:NLR7301mesh} for an illustration). 
The Reynolds number is fixed in all cases at $\text{Re}=1.7\times 10^{6}$ and two independent operating parameters are varied across simulations: the Mach number $\mathscr{M}\in[0.3,\,0.75]$ and the angle of attack $\alpha \in[-3^{\circ},\,5^{\circ}]$.
%From the initial 100 candidate operating points, generated using a Halton sequence, $m=98$ CFD solutions remain after discarding non-converged runs. 
The dataset $\mathscr{S}$ \cite{hines2023graph} comprises a total of $m=98$ different CFD simulations of the RANS equations obtained by varying $(\mathscr{M}, \alpha)$ so as to adequately cover the $\mathscr{M}-\alpha$ plane.
Accordingly, $\mathscr{S} = \{(z_k, y_k)\}_{k=1}^m$, where the input vector of the $k$-th datapoint $z_k=(\mathscr{M}, \alpha)_k$ determines the operating condition, and its output vector $y_k=[C_p(x_1), C_p(x_2),\dots,C_p(x_n)]_k$ determines the pressure coefficients $C_p$ at each of the $i=1,2,\dots,n$ airfoil's boundary nodes.\\
Following \cite{hines2023graph}, $\mathscr{S}$ is then adequately split into training, validation and test sets $\mathscr{S} = \mathscr{S}_{\text{training}} \cup \mathscr{S}_{\text{validation}} \cup \mathscr{S}_{\text{test}}$, with $m_{\text{training}}=39$, $m_{\text{validation}}=20$, and $m_{\text{test}}=39$, see \autoref{fig:nlr7301_dataset} for an illustration.

\subsection{Surrogate models}
Three different surrogate models --a MLP, a GNN and a KAN, see below-- are all initially trained on $\mathscr{S}_{\text{training}}$. Both the MLP and KAN-based surrogate models are here considered as {\it local predictors}, i.e. at inference both these models are given a pair $(\mathscr{M}, \alpha)$ {\it and a specific node location $x_i$} as inputs, and these models yield the prediction of the pressure coefficient at that specific location $$\hat{C_p}^{\text{MLP}}=f^{\text{MLP}}(\mathscr{M},\alpha, x_i) \in \mathbb{R}; \ \ \hat{C_p}^{\text{KAN}}=f^{\text{KAN}}(\mathscr{M},\alpha, x_i) \in \mathbb{R}, \forall i=1,2,\dots,n.$$
In other words, both MLP and KAN will take a triplet $(z,x)$ as the input vector. 
On the other hand, the GNN-based surrogate model is a {\it global predictor}, such that, at inference, this model only requires specification of the pair $(\mathscr{M}, \alpha)$, and the model predicts the pressure coefficient at all the $n$ spatial locations at once:
$$\hat{C_p}^{\text{GNN}}=f^{\text{GNN}}(\mathscr{M},\alpha); \ \ \ \hat{C_p}^{\text{GNN}} = [\hat{C_p}(x_1),\hat{C_p}(x_2),\dots, \hat{C_p}(x_n)] \in \mathbb{R}^n.$$
For convenience, in what follows we will use $z\equiv (\mathscr{M}, \alpha)$ or $z\equiv (\mathscr{M}, \alpha,x)$ depending on the context.\\
Below we provide a reminder on the basic properties of the MLP, GNN and KAN architectures. In order to compare our baseline results on MLP and GNN with previous works, our selection of the MLP and GNN specification matches the one in \cite{hines2023graph}.

\medskip\noindent {\bf MLP --} 
Multi-Layer Perceptrons (MLPs) are the foundation of many deep supervised learning models, thanks to their expressivity as universal approximators. 
A multilayer perceptron (MLP) with $L$ hidden layers defines a nonlinear mapping from an input vector (in our case $z=(\mathscr{M},\alpha,x) \in \mathbb{R}^{3}$, where $x$ defines a spatial node location in the airfoil boundary) to an output vector (in our case $\hat{y}=\hat{C_p}(x) \in \mathbb{R}$) through a sequence of affine transformations followed by element-wise nonlinear activation functions.\\
Formally, denoting the hidden-layer activations by $\mathbf{h}^{(l)}$, the MLP is a nonlinear function that can be written recursively as\\
$\mathbf{h}^{(l)} = \sigma\!\left(\mathbf{W}^{(l)}\mathbf{h}^{(l-1)} + \mathbf{b}^{(l)}\right)$ for $l=1,\dots,L$, with $\mathbf{h}^{(0)}=z$, where $\mathbf{W}^{(l)}$ and $\mathbf{b}^{(l)}$ are the weight matrices and bias vectors, and $\sigma(\cdot)$ is a scalar nonlinear activation function, which in our case is the exponential
linear unit 
$$
\mathrm{ELU}(w) =
\begin{cases}
w, & w > 0, \\
e^{w} - 1, & w \le 0.
\end{cases}
$$
The optimal number of layers $L$ is an hyperparameter found after an hyperparameter optimisation, as is explained in the next subsection.\\
The optimal assignment of network parameters $\theta=\{\mathbf{W}^{(l)},\mathbf{b}^{(l)}\}_{l=1}^{L+1}$ are learned by minimizing a loss function $\mathcal{L}(\theta)$ over the training dataset $\mathscr{S}_{\text{training}}$ via an Adam optimiser. In our case, we use a mean-square error (MSE) type, where
\begin{equation}
    \mathcal{L}(\theta)=\frac{1}{m_{\text{training}}}\sum_{k=1}^{m_{\text{training}}}\bigg(\frac{1}{n}\sum_{i=1}^{n}[\hat{C_p}(x_i,z_k)-C_p(x_i,z_k)]^2\bigg).
    \label{eq:loss}
\end{equation}
where $\hat{C_p}(x_i,z_k), \ C_p(x_i,z_k)$ are respectively the predicted pressure coefficient for operating conditions $z_k$ evaluated at the spatial node $x_i$, and its ground true value (the one provided by the RANS simulation) at each location of the airfoil boundary. 

\medskip\noindent {\bf GNN --} A Graph Neural Network (GNN) \cite{scarselli2008graph, corso2024graph} extends the multilayer perceptron (MLP) paradigm to input data that are naturally defined on (or benefit from) graphs, instead of vectors. Accordingly, while an MLP applies a sequence of dense affine transformations to a fixed input vector, a GNN operates on a graph $G=(V,E)$ with node features $\mathbf{h}_i^{(0)}$, and updates each node representation by aggregating information from its neighbors. At layer $l$, the representation of node $i$ is typically updated as
\[
\mathbf{h}_i^{(l)} = \sigma\!\left( W^{(l)} \, \text{AGG}\big(\{\mathbf{h}_j^{(l-1)} : j \in \mathcal{N}(i)\}\big) \right),
\]
where $\mathcal{N}(i)$ denotes the neighbors of node $i$, AGG is a permutation-invariant aggregation operator (e.g., sum, mean, or max), $W^{(l)}$ is a learnable weight matrix, and $\sigma$ is a nonlinear activation function. In contrast to an MLP, which treats inputs as independent features, the GNN explicitly propagates information along graph edges, allowing the learned representation of each node to depend on the features of its local neighborhood. 
\\
When applied to an airfoil, a Graph Neural Network (GNN) operates on a graph whose nodes correspond to discrete points along the airfoil surface, typically ordered along the contour so that the graph forms a closed chain. Each node carries local features such as its coordinates or other geometric descriptors of the airfoil. Edges connect neighboring surface nodes, reflecting the physical adjacency of points along the airfoil. During each GNN layer, the feature vector at a node is updated by aggregating information from its neighboring nodes and combining it with its own features through a learnable transformation and nonlinearity. This structure is particularly suitable for modeling aerodynamic quantities such as pressure coefficients, which are not independent at each surface point: due to the continuity and smoothness of the flow field along the surface, the pressure at a given location is strongly correlated with the pressure at nearby locations. By propagating information along the graph, the GNN can naturally learn and exploit these local correlations while still capturing the influence of the global airfoil geometry. The loss function matches Eq.~\ref{eq:loss}, and training is also done via Adam optimisation.\\
Unlike MLP, for GNN some additional geometric information is needed as input features such as surface normals and face normals, see \cite{hines2023graph} for details. 
%are expressed in polar coordinates. Ultimately, min-max scaling is performed independently on each node and edge feature utilizing the statistics derived from the training data.ESTE ULTIMO PARRAFO ES UN POCO CONFUSO, HABRIA QUE ACLARAR MAS LO DE LA GNN O QUITAR ESTO Y QUE LA GENTE SE LEA EL PAPER DE DLR PARA MAS DETALLE.}

% \begin{table}[H]
%    \centering
%    \begin{tabular}{|l|l|l|l|}
%        \hline
%        \textbf{GNN Features} &  \\
%        \hline
%        Node &  $\mathscr{M}, \alpha, x_i, \nu_i$\\
%        Edge & $\mathbf{d}_{ji}, \mathbf{f}_{ji}$ \\
%        \hline
%    \end{tabular}
%    \caption{Summary of input features for the GNN model.}
%    \label{tab:summary}
%\end{table}

\medskip\noindent {\bf KAN --} Both MLPs and GNNs rely on the their expressivity as universal approximators, as per the universal approximation theorem \cite{hornik1991approximation}. In turn, the Kolmogorov-Arnold representation theorem \cite{kolmogorov1961representation} states that any continuous multivariate function can be represented as a {\it finite} sum of compositions of univariate functions. Of course, the challenge lies in finding the precise shape of these univariate functions. Very recently, Liu and collaborators introduced Kolmogorov--Arnold Networks (KANs) \cite{liu2024kan, pykan_repositorio} along with a practical way to {\it learn} those univariate functions from data (albeit chosen within a finite basis of candidate functions --initially a spline basis--, hence representation is not, in general, exact). Accordingly, KANs build a nonlinear mapping by replacing the (learnable) scalar weights of a standard MLP with learnable (parametric) univariate functions on the edges of the neural network. For our input vector $z \in \mathbb{R}^3$, each layer computes a sum of transformed inputs, where the transformations are parameterised one–dimensional functions. Denoting the activations at layer $l$ by $\mathbf{h}^{(l)}$, a KAN layer can be written as
\[
h^{(l)}_j = \sum_{i} \phi^{(l)}_{ij}\!\left(h^{(l-1)}_i\right), \quad l=1,\dots,L,
\]
with $\mathbf{h}^{(0)}=z=(\mathscr{M}, \alpha, x)$, where $\phi^{(l)}_{ij}(\cdot)$ are learnable univariate functions (represented using spline bases in the seminal work \cite{pykan_repositorio}, but here we use Chebyshev polynomials as the orthonormal basis). The network output is obtained from the final layer in the same additive form. The parameters defining these functions are learned again by minimizing a MSE-type loss similar 
%\textcolor{magenta}{SIMILAR OR SAME?}\textcolor{blue}{LL -- he puesto similar porque los parámetros a optimizar son distintos simplemente, pero que los chicos confirmen que esa es la única diferencia} 
to the case of the MLP over $\mathscr{S}_{\text{training}}$.

\medskip Each of the three models have an additional set of parameters which are not trained via loss minimization (e.g. the number of layers, some parameters of the optimiser, etc). For reproducibility, in the case of the MLP and the GNN surrogates, these hyperparameters are initially chosen following the hyperparameter optimisation in \cite{hines2023graph}, and are depicted in \autoref{tab:hyperparams_MLP} and \autoref{tab:hyperparams_GNN} for the MLP and the GNN, respectively. The hyperparameter optimisation of the KAN model is detailed in Sec.~\ref{subsec:optuna}. All three models are trained on the training set $\mathscr{S}_{\text{training}}$. For hyperparameter optimisation, we train models with different configurations of hyperparameters in $\mathscr{S}_{\text{training}}$ and select as the best configuration the one for which the model shows smaller MSE in the validation set $\mathscr{S}_{\text{validation}}$. The generalisation performance of the best models is then assessed in the test set $\mathscr{S}_{\text{test}}$, see below for details.

\begin{table}[H]
    \centering
    \begin{tabular}{@{}ll@{}}
        \toprule
        \textbf{Hyperparameter} & \textbf{Value} \\ 
        \midrule
        Initial learning rate & $3.45 \times 10^{-4}$ \\
        LR decay factor & $0.9954$ \\
        Batch size & $415$ \\
        Dimension of hidden layers & $256$ \\
        Number of hidden layers & $11$ \\
        \midrule
        Trainable parameters & $725249$ \\
        \bottomrule
    \end{tabular}
    \caption{Hyperparameters used for the MLP model, taken from \cite{hines2023graph}.}
    \label{tab:hyperparams_MLP}
\end{table}

\begin{table}[H]
    \centering
    \begin{tabular}{@{}ll@{}}
        \toprule
        \textbf{Hyperparameter} & \textbf{Value} \\ 
        \midrule
        Initial learning rate & $6.50 \times 10^{-4}$ \\
        LR decay factor & $0.9954$ \\
        Parameter batch size & $15$ \\
        Node batch size & $32$ \\
        Dimension of hidden layers & $256$ \\
        Latent dimension & $16$ \\
        Encoder hidden layers & $6$ \\
        GNN layers & $1$ \\
        $MLP_{M_t}$ hidden layers & $2$ \\
        $MLP_{U_t}$ hidden layers & $2$ \\
        Decoder hidden layers & $1$ \\
        \midrule
        Trainable parameters & $299285$ \\
        \bottomrule
    \end{tabular}
    \caption{Hyperparameters used for the GNN model, taken from \cite{hines2023graph}.}
    \label{tab:hyperparams_GNN}
\end{table}

\noindent {\bf Error quantification --} Once the three models are properly trained, their generalisation peformance is initially tested on $\mathscr{S}_{\text{test}}$ using a variety of standard global quantifiers \cite{Goodfellow-et-al-2016,kuhn2013applied}. Instead of a MSE-type error (Eq.~\ref{eq:loss}), we use the mean absolute error (MAE) and the root mean squared error (RMSE), where the latter one is essentially the square root of Eq.~\ref{eq:loss} (computed on the test set), whereas the former is like Eq.~\ref{eq:loss} but using the absolute value of the residuals, instead of the square value. Additionally, from the population of all absolute values of the residuals $\{|\hat{C_p}(x_i,z_k)-C_p(x_i,z_k)|\}$ we extract their 95 and 99 percentiles, as indicators of extreme errors. Finally, we also compute the coefficient of determination $R^2$, computed from a scatter plot between the ground true pressure coefficient and the prediction, for all spatial locations and operating conditions in the test set.

%A set of standard error metrics is used to assess and compare the overall performance of the models (vid. Table \ref{tab:err_metrics}). Given the one-dimensional nature of the dataset and the widespread adoption of these metrics in machine learning \cite{Goodfellow-et-al-2016,kuhn2013applied}, this selection provides a simple and effective basis for evaluation and comparison with previous work.

%\begin{table}[H]
%    \centering
%    \renewcommand{\arraystretch}{3}
%    \begin{tabular}{|c|c|}
%        \hline
%        \textbf{Error metric} & \textbf{Formula} \\
%        \hline
%        Mean Absolute Error (MAE) & $\displaystyle \frac{1}{n} \sum_{i=1}^{n} |\mathscr{C}_p(x_i) - \hat{\mathscr{C}_p}(x_i)|$ \\
%        Root Mean Squared Error (RMSE) & $\displaystyle \sqrt{\frac{1}{n} \sum_{i=1}^{n} [\mathscr{C}_p(x_i) - \hat{\mathscr{C}_p}(x_i)]^2}$ \\
%        Absolute Error Percentile 95 (AEq95) & $\displaystyle \text{Percentile}_{95}(|\mathscr{C}_p(x_i) - \hat{\mathscr{C}_p}(x_i)|)$ \\
%        Absolute Error Percentile 99 (AEq99) & $\displaystyle \text{Percentile}_{99}(|\mathscr{C}_p(x_i) - \hat{\mathscr{C}_p}(x_i)|)$ \\
%        Determination coeff. ($R^2$) & $\displaystyle 1 - \frac{\sum_{i=1}^{n} [\mathscr{C}_p(x_i) - \hat{\mathscr{C}_p}(x_i)]^2}{\sum_{i=1}^{n} [\mathscr{C}_p(x_i) - \bar{\mathscr{C}_p}]^2}$ \\
%        \hline
%    \end{tabular}
%    \caption{Error metrics.}
%    \label{tab:err_metrics}
%\end{table}

In addition, observe that spatial correlations are expected to emerge among the set of surface pressure coefficients, due to physical constraints and simple continuity arguments. Accordingly, those correlations should also be captured by the surrogate model, even if it is a local regressor. To that aim, for each model we construct two $597\times 597$ correlation matrices \(\mathscr{C}^{\text{true}}\) and \(\mathscr{C}^{\text{pred}}\). The $ij$-entry of these matrices denote, respectively, the (properly normalised) correlation between the pressure coefficient at spatial location $x_i$ and the one at spatial location $x_j$ (computed over all the operating conditions $z_k$ in the test set) for the ground truth and the model predictions, i.e.  
\begin{equation}
\mathscr{C}^{\text{true}}_{ij} = \frac{1}{m_{\text{test}}}\sum_{k=1}^{m_{\text{test}}}\frac{[C_p(x_i,z_k) - \mu_i][C_p(x_j,z_k) - \mu_j]}{\sigma_i\sigma_j},
\label{eq:corr}
\end{equation}
where $\mu_X$ and $\sigma_X$ are the average and standard deviation of all the pressure coefficients at spatial location $X$ (over the ensemble of operating conditions considered in $\mathscr{S}_{\text{test}}$).
%Comparison of \(\mathscr{C}^{\text{true}}\) vs \(\mathscr{C}^{\text{pred}}\) across our three surrogate models allows us to assess (i) what regions of the airflow 
%that reproduces the true correlation structure is expected to capture global flow features and preserve the relationships inherent in the data. 
Finally, to quantify the discrepancy between \(\mathscr{C}_{\text{pred}}\) and \(\mathscr{C}_{\text{true}}\) for each model we use the Frobenius relative error \parencite{golub2013matrix}:
\begin{equation}
    \varepsilon_F = \frac{\|\mathscr{C}_{\text{pred}} - \mathscr{C}_{\text{true}}\|_F}{\|\mathscr{C}_{\text{true}}\|_F}.\label{eq:frob}
\end{equation}
Lower values of \(\varepsilon_F\) indicate a closer match to the true correlation structure.

\section{Results}
\label{sec:results}
\subsection{Hyperparameter optimisation and impact of RNG seed: KAN vs MLP}\label{subsec:optuna}
At odds with the MLP and the GNN architectures \cite{hines2023graph}, KAN-based surrogate models have not, to the best of our knowledge, been tested before in the context of this learning task. Accordingly, for this model we need to perform an adequate hyperparameter optimisation in $\mathscr{S}_{\text{validation}}$. In addition, this section also explores the possible dependence on the performance of both KAN and MLP models with respect to the seed of the pseudo random number generator (RNG), which impacts (i) the weight initialization of the networks, and (ii) the selection of mini-batches. Such potential dependence is seldom explored in the literature and, as we shall see, its impact for the KAN performance is (unexpectedly) substantially larger than for the MLP.

\subsubsection{Hyperparameter optimisation for the KAN model}
\begin{figure}[htb!]
    \centering
    \includegraphics[width=0.95\linewidth]{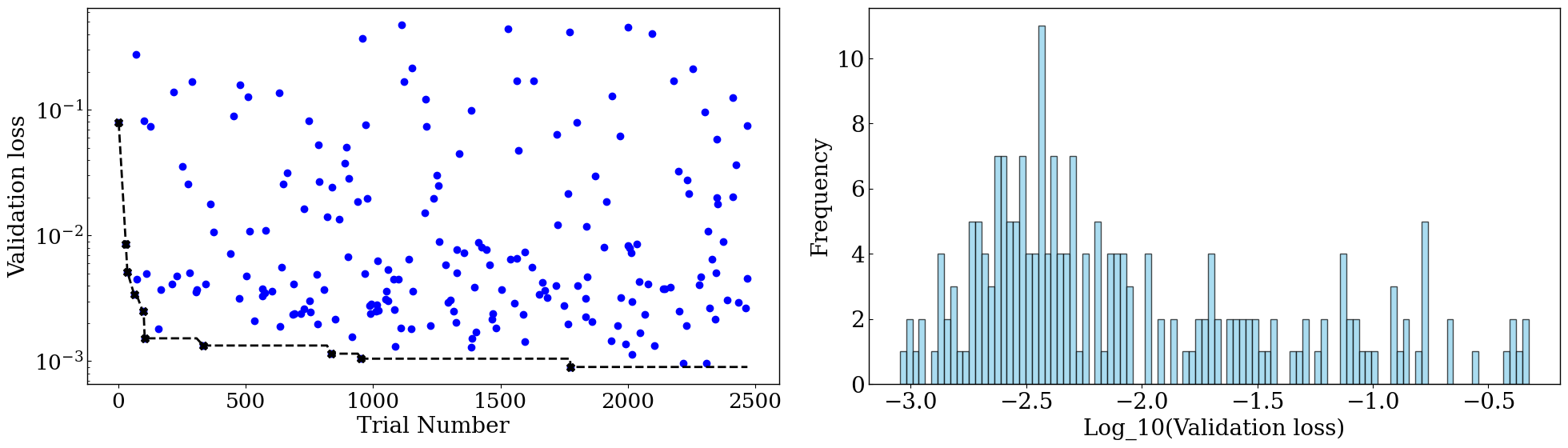}
    \caption{Performance of trained KAN-based models (in terms of MSE loss in the validation set) for a quasi-Montecarlo sampling of hyperparameters. ({\it Left}) The left panel reports the MSE loss for the different trials of hyperparameters (the black dots depict how the smallest detected MSE changes as trials progress). Observe that we initially check 2472 trials, but only 312 produced converged resuls (blue dots). ({\it Right}) Linear-log frequency histogram of MSE loss (in the validation set) for the 312 sets of hyperparmaters. Observe that the histogram has a long right tail, suggesting that the performance of the KAN-based model is highly dependent on its hyperparameters.}
    \label{fig:qmc_study_process_hist}
\end{figure}

The set of hyperparameters include the number of layers and number of neurons per layer, the learning rate decay factor in the Adam optimiser, or the degree of the Chebyshev polynomials in the trainable activation functions, among others (see \autoref{tab:hyperparams_KAN} for details).
After selecting the initial range for each of the hyperparameters, we perform two complementary analysis.\\
In the first one, we sample at random (quasi-Montecarlo) a total of 2472 different hyperparameter configurations. For each configuration, we train the KAN-based model on the training set $\mathscr{S}_{\text{training}}$, and compute its MSE-based loss on the validation set $\mathscr{S}_{\text{validation}}$. Among the initial 2472 different hyperparameter configurations, the KAN-based model only converged in 312 cases, and the rest were discarded mostly due to exploding gradients. This feature highlight the intrinsic instability of the KAN architecture with respect to hyperparameters \cite{shukla2024comprehensive}. Results for the remaining 312 trials is reported in \autoref{fig:qmc_study_process_hist}. Observe that the distribution of MSE losses is long-tailed (note that the frequency histogram is in linear-log scales), highlighting that whereas a large number of configurations have small validation MSE loss, a non-negligible amount of configurations substantially deviate and can attain much larger errors, i.e. the dependency of the KAN-based model's performance on the selection of its hyperparameters is strong, with many poorly-performing outliers emerging if the hyperparameters are not carefully chosen.

\begin{figure}[H]
    \centering
    \begin{subfigure}{0.49\linewidth}
        \centering
        \includegraphics[height=7.5cm]{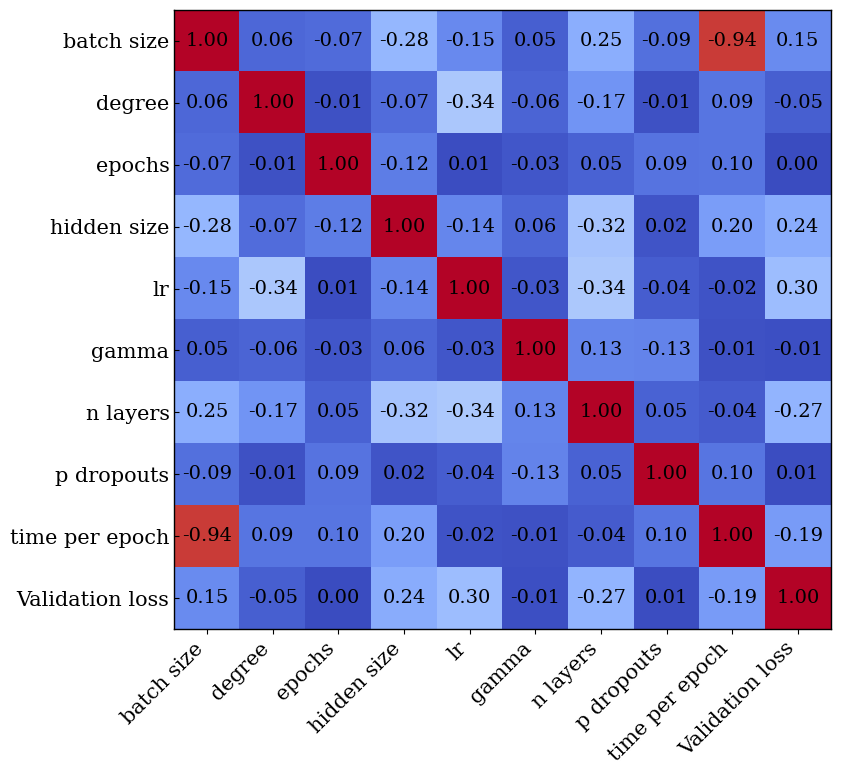} %width=\linewidth
        \caption{Spearman matrix.}
        \label{fig:spearman_matrix_KAN_repeated}
    \end{subfigure}\hfill
    \begin{subfigure}{0.49\linewidth}
        \centering
        \includegraphics[height=7.5cm]{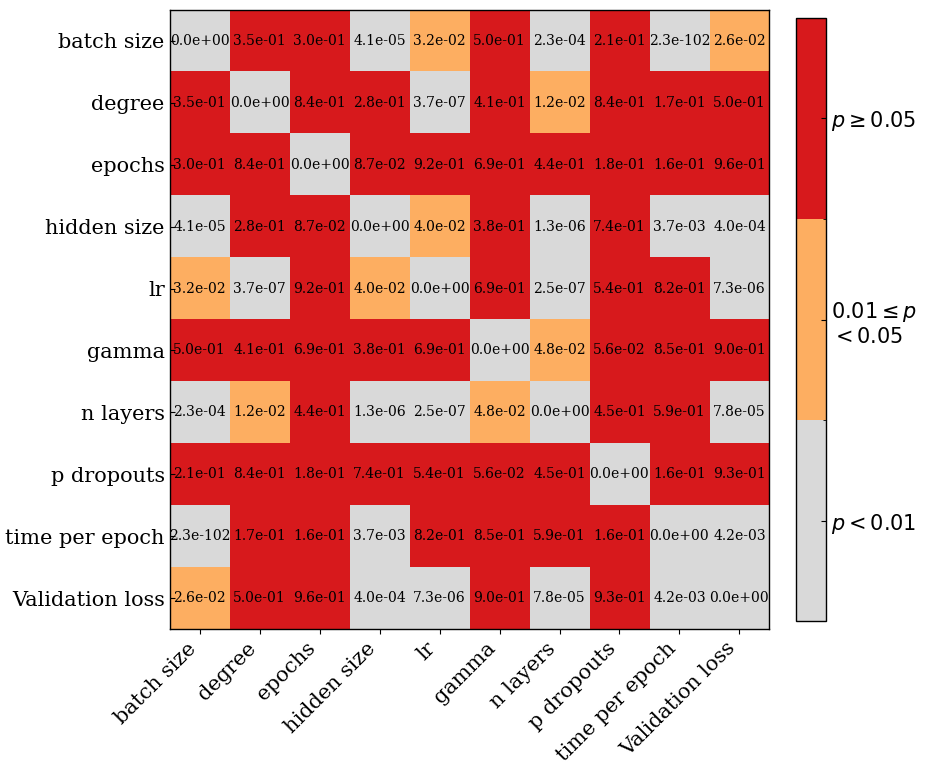} %width=\linewidth
        \caption{p-values matrix.}
        \label{fig:Spearman_pvalues_repeated}
    \end{subfigure}
    \caption{Spearman correlation analysis between all pairs of hyperparameters (and the validation loss) computed over the 312 hyperparameter configurations (those surviving from the initial 2472 trials).}
    \label{fig:kan_results}
\end{figure}

\medskip \noindent 
We then investigate whether tendencies  emerge between increasing/decreasing values of the hyperparameters and performance in the validation set, by computing Spearman correlation coefficients (and their statistical significance), see \autoref{fig:Spearman_pvalues_repeated}. The correlation between selected pairs of hyperparameters is generally low, although some for some pairs such correlation appear to be statistically significant (gray entries in the p-value matrix). The last row of these matrices indicate the dependence on the loss of different hyperparameters, showing moderate yet significant tendency for the learning rate (larger learning rates lead to larger loss error) and the number of layers (a larger number moderately correlate with a smaller error). These dependencies are overall weak and non-informative, see for instance table \ref{tab:hyperparams_KAN} where the optimal number of layers is found to be small.

\begin{table}[H]
    \centering
    \begin{tabular}{@{}llll@{}}
        \toprule
        \textbf{Hyperparameter} & \textbf{Minimum value} & \textbf{Maximum value} & \textbf{Optimal value} \\ 
        \midrule
        Number of hidden layers  & $1$ & $6$ & $2$ \\ 
        Neurons per layer        & $10$ & $400$ & $25$ \\ 
        Batch size ($2^p$)       & $p = 4$ & $p = 10$ & $p = 7$ \\ 
        Learning rate decay factor ($\gamma$) & $0.9$ & $0.999$ & $9.5071 \times 10^{-1}$ \\ 
        Degree of Chebyshev polynomials & $4$ & $8$ & $4$ \\ 
        Dropout probability      & $10^{-5}$ & $10^{-1}$ & $1.7878 \times 10^{-5}$ \\ 
        Number of epochs         & $100$ & $800$ & $474$ \\ 
        Initial learning rate    & $10^{-5}$ & $10^{-2}$ & $1.3024 \times 10^{-3}$ \\  
        \midrule
        Trainable parameters & & & $6875$ \\
        \bottomrule
    \end{tabular}
    \caption{Range of the different hyperparameters and their optimal configuration obtained for the KAN model via Optuna.}
    \label{tab:hyperparams_KAN}
\end{table}

\begin{figure}[htb!]
    \centering
    \includegraphics[width=0.95\linewidth]{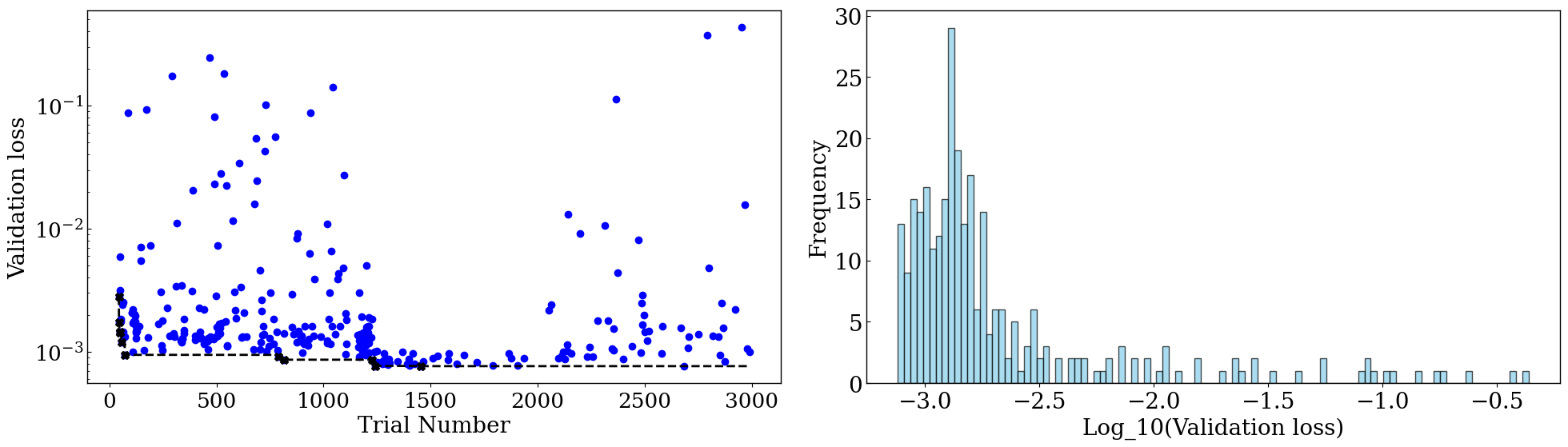}
    \caption{Same as \autoref{fig:qmc_study_process_hist} but applied to a selection of hyperparameters from Optuna \cite{optuna_2019}, that uses a Tree-Structured Parzen Estimator (TPE). The tail of large MSE loss in the frequency histogram (right panel) has less mass, but is not completely removed.}
    \label{fig:optimal_study_process_hist}
\end{figure}

\medskip \noindent 
The second analysis repeats this whole procedure albeit performing a non-uniform sampling of the hyperparameters with the Optuna Python library \cite{optuna_2019}, concretely using a Tree-Structured Parzen Estimator (TPE). A total of 311 converged trials (hyperparameter configurations) are found, and the frequency histogram of resulting MSE losses in the validation set for this hyperparameter sampling is reported in \autoref{fig:optimal_study_process_hist}. Compared to \autoref{fig:qmc_study_process_hist}, we observe that this non-uniform sampling concentrates the measure in the region of small MSE losses, although the long tail is not completely removed. We use this sampling to select the hyperparameter configuration that minimises the MSE on the validation set. Interestintly, the optimal configuration (the left end of the histogram) itself is not an outlier among the distribution of configurations for which the model is properly trained. This suggests that the performance of this optimal KAN model is reasonably robust against small changes in hyperparameters. The optimal configuration is depicted in \autoref{tab:hyperparams_KAN}.\\

%\textcolor{orange}{Se ha hecho un primer estudio de Optuna con el objetivo de encontrar una correlación de los parámetros seleccionados con el error en el set de validación, utilizando un muestreo basado en un algoritmo de Monte Carlo (https://docs.scipy.org/doc/scipy/reference/stats.qmc.html). Posteriormente se ha realizado un siguiente estudio de Optuna para buscar la configuración óptima del modelo en el caso a estudiar, minimizando el error en el set de validación. Por último se ha estudiado la relevancia de la semilla de los Generadores de Números Random (RNG) en las métricas del error alcanzado, utilizando esta semilla como un hiperparámetro más y analizando la dispersión que genera.}
%\textcolor{orange}{En el primer estudio (figura \ref{fig:qmc_study_process_hist}, se han completado 312 trials y se puede ver la variabilidad del error en validación en la figura \ref{fig:qmc_study_process_hist}. Una primera conclusión que se desprende del histograma de la figura \ref{fig:qmc_study_process_hist} es la disparidad entre valores posibles que da el modelo en función del valor de los hiperparámetros. Esta afirmación suma importancia a la necesidad de una optimización de hiperparámetros para encontrar un modelo válido en la aplicación deseada.}

%\textcolor{orange}{Para asegurar la falta de sesgo por parte de Optuna en la elección de los valores de hiperparámetros, se hace alusión a los colores de la figura \ref{fig:spearman_matrix_KAN}, que muestra el valor del coeficiente de Spearman de cada uno con respecto a los demás hiperparámetros y al error en validación. La tonalidad azúl de la mayoría de ellos (valor absoluto del coeficiente de Spearman cercano a cero) indican, en primera aproximación, que no existe a priori relación entre los hiperparámetros escogidos, y que por tanto no hay un sesgo en el espacio de estudio. La única relación que se puede sacar con suficiente certeza es la dependencia inversa entre el tiempo de cálculo por época y el tamaño del batch, naturaleza que ya se conocía de antemano al estar directamente relacionado con el número de veces que se ejecuta el algoritmo.}
%\textcolor{orange}{La última fila de las figuras \ref{fig:spearman_matrix_KAN_repeated} y \ref{fig:Spearman_pvalues_repeated} corresponde a la correlación posible entre todos los hiperparámetros y el error en validación esperado. En general, no se observan tendencias significativamente importantes, ya que todos los valores estén cercanos a cero y algunos ellos incluso tengan un p-value menor a $0.05$.}

%\textcolor{orange}{El segundo estudio realizado (ver figura \ref{fig:optimal_study_process_hist}), compuesto por 311 trials, minimiza el error de validación permitiendo a Optuna muestrear el espacio de hiperparámetros con tal fin. De este proceso se saca la configuración óptima del modelo que se presentará más adelante como el definitivo. }
%\textcolor{orange}{Tanto la configuración óptima como el espacio de diseño de los hiperparámetros están reflejados en la tabla \ref{tab:hyperparams_KAN}. Sin mucho más que añadir, ya se comparará con los otros modelos más adelante.}

\subsubsection{Impact of the RNG seed}
\begin{figure}[H]
    \centering
    \begin{subfigure}{0.45\linewidth}
        \centering
        \includegraphics[height=4.5cm]{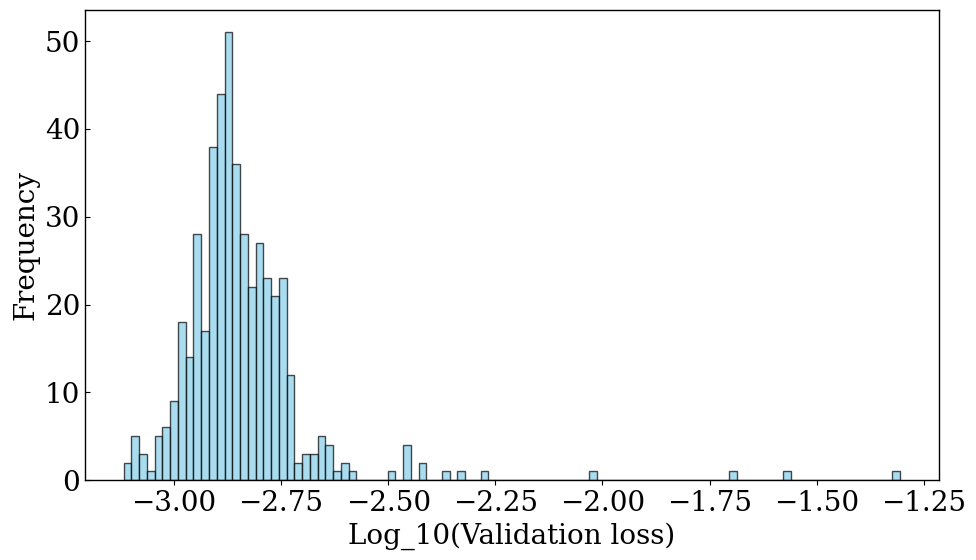} %width=\linewidth
        \caption{KAN (468 different seed trials).}
        \label{fig:kan_seed_study_process_hist}
    \end{subfigure}
    \hfill
    \begin{subfigure}{0.45\linewidth}
        \centering
        \includegraphics[height=4.5cm]{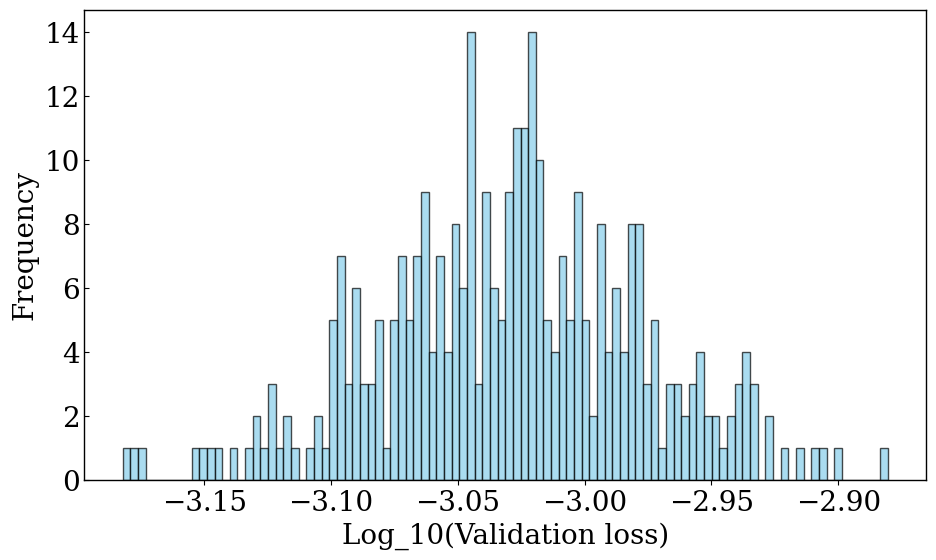} %width=\linewidth
        \caption{MLP (334 different seed trials).}
        \label{fig:mlp_seed_study_process_hist}
    \end{subfigure}
    \caption{Linear-log frequency histograms of MSE loss (in the validation set) for the KAN-based model (panel a) and the MLP model (panel b) obtained when uniformly samply different RNG seeds. The total number of different sampled seeds is 468 for the KAN model and 334 for the MLP model.
    The impact of choosing different RNG seeds has a substantially larger impact for the KAN-based model, where we find seed instances that lead to outlier results in terms of very large MSE loss, whereas in the MLP model performance is relatively robust against seed variation.}
    \label{fig:seed_studies}
\end{figure}

As an additional analysis, we explore --both for the KAN and MLP-based models-- the impact that different RNG seeds might have on the model's performance in the validation set. Observe that changing the seed of the pseudo-random number generator changes (i) the model's initialisation, and (ii) the order in which the data batches enter into the optimisation loop within each epoch.
Different seeds are sampled uniformly at random and without replacement (a total of 468 seeds are considered for the KAN-based model, and 280 seeds for the MLP).
Results are shown in panels (a) and (b) of \autoref{fig:seed_studies}. While the MLP model is relatively robust against changing the RNG seed as expected \cite{picard2021torch}, we find that in the KAN model there are some RNG seeds that induce the model to perform poorly at validation set. This reinforces the evidence presented in Figs.~\ref{fig:qmc_study_process_hist} and \ref{fig:optimal_study_process_hist} about the strong dependence of KAN-based model's performance on hyperparameters. 

%\textcolor{orange}{Por último, se ha estudiado la influencia de la semilla del Generador de Números Random (RNG) en los resultados del modelo. Esto afecta íntegramente a la inicialización de los pesos de la red, así como en el orden en que se entregan los batch en el entrenamiento. Como estrategia de muestreo, se han cogido valores de semilla en distribución uniforme y sin posibilidad de repetir. Se ha realizado tanto para la KAN como para el MLP y se representa en las figuras 
%\ref{fig:kan_seed_study_process_hist} y \ref{fig:mlp_seed_study_process_hist}. }

\subsection{Learning curves}
\begin{figure}[htb!]
    \centering
    \includegraphics[width=0.55\linewidth]{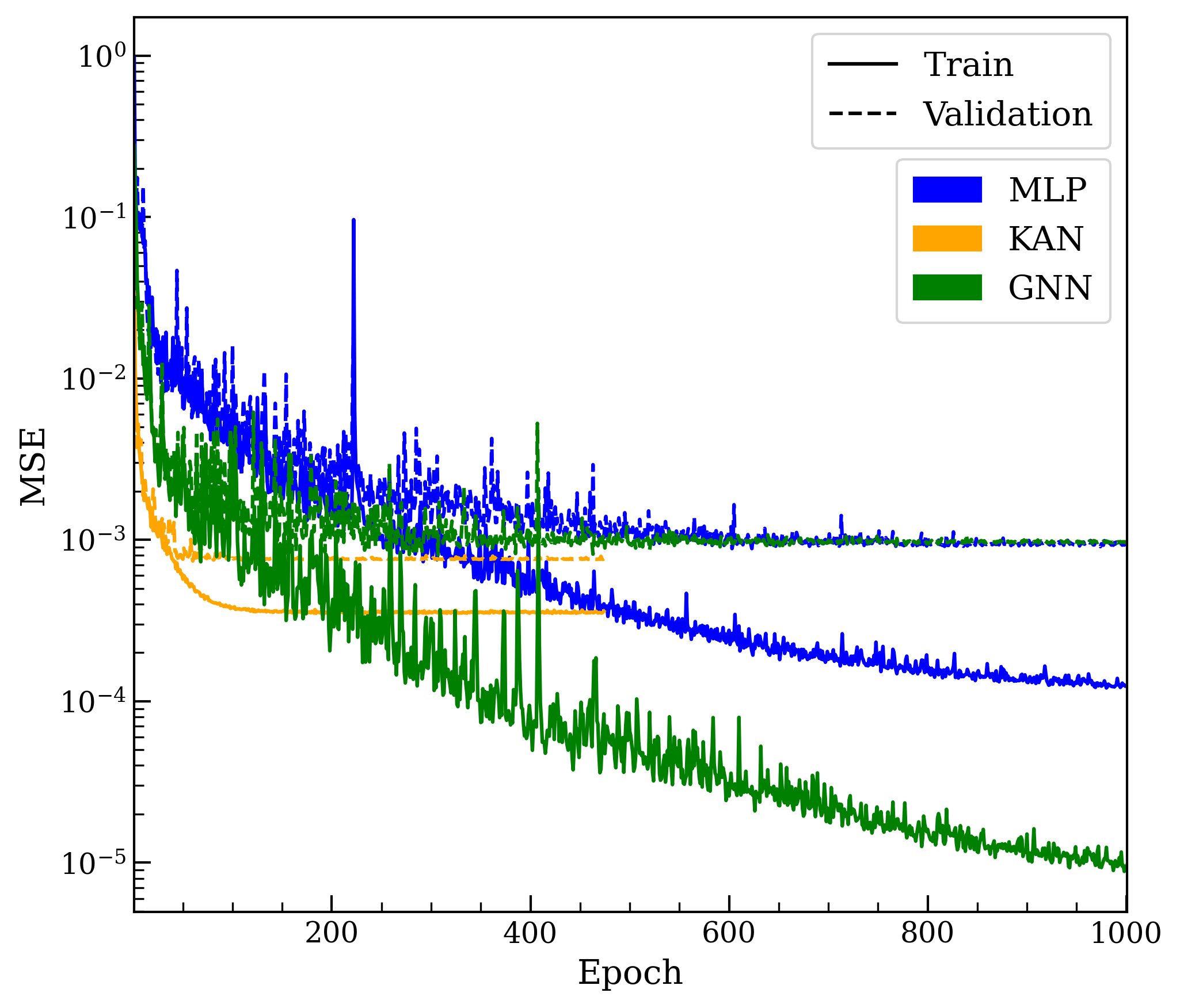}
    \caption{Training and validation MSE loss curves as a function of the number of epochs for the MLP, KAN, and GNN models after hyperparameter optimization.}
    \label{fig:training_curves_modelos}
\end{figure}

Once the hyperparameters of the models have been chosen, a final training run is carried out using the best configurations. The corresponding learning curves (semi-log plots of MSE-based loss as a function of the training progress, calculated both in the training --solid lines-- and validation sets --dashed lines--) for the three models are presented in \autoref{fig:training_curves_modelos}. Both the MLP and GNN models are trained over 1000 epochs, whereas the KAN model requires smaller number of epochs (hyperparameter optimisation yields roughly below 500 epochs). 
%\textcolor{orange}{FG: en la figura ya no representamos el número de epochs en el eje x, ahora hemos puesto el training progress como porcentaje. Esto es porque al final la KAN requería menos epochs que MLP y GNN. Hemos entrenado 474 epochs para la KAN, mientras que MLP y GNN 1000 epochs. No sé si deberíamos hacer algún comentario sobre esto en el texto o no}. 
On the training set, both MLP and  GNN exhibit comparable learning behavior (with the GNN achieving lower training losses), and continue to decrease the training loss even after 1000 epochs. Comparing these against the curves on the validation set, which converge after around 500 epochs, suggest that for longer epochs both MLP and GNN are overfitting to the training set, with no impact on validation. MLP and GNN attain comparable loss performance on the validation set, stabilising at a MSE loss around $10^{-3}$. The KAN model in turn seems to learn much faster, with both training and validation curves stabilising after about 50 epochs --i.e. only 10\% of the number of epochs needed for both MLP and GNN--, achieving a MSE loss in the validation set of $8\times 10^{-4}$, i.e. slightly lower than the MLP and GNN models.

%The KAN model in turn seems to learn much faster (both training and validation curves stabilise after about 50 epochs), although the attained MSE loss in the validation set is slightly higher $\approx 2\times 10^{-3}$.
% \textcolor{red}{COMPROBAR TODOS LOS NUMEROS, QUE CONCUERDEN CON LAS EXPLICACIONES Y EL RESTO DE GRAFICAS, CUIDADO CON TEMAS DE RESCALADO ETC}
%\textcolor{red}{EN LA FIGURA SOLO HAY TRAIN Y VALIDATION, NO HAY TEST, HAY Q CAMBIAR LA LEYENDA DE LA FIGURA?}

%However, both models ultimately reach comparable levels of generalization performance in the validation set. In contrast, the KAN displays a markedly different trend: although its loss decreases rapidly during the initial epochs, the learning process halts around epoch 50, with no significant improvements thereafter. As a result, both the training and validation losses of the KAN remain consistently higher than those of the MLP and GNN models.

\begin{table}[htb!]
    \centering
    \begin{tabular}{lccccc}
        \toprule 
        & \multicolumn{2}{c}{\textbf{DLR \cite{hines2023graph}}} & \multicolumn{3}{c}{\text{this work}} \\
        \cmidrule(lr){2-3} \cmidrule(lr){4-6}
        & \textbf{MLP} & \textbf{GNN} & \textbf{MLP} & \textbf{GNN} & \textbf{KAN} \\
        \midrule
                \textit{MSE}   &          &          & $9.7 \times10^{-4}$ & $8.4 \times 10^{-4}$ & $1.2\times10^{-3}$ \\
        \textit{MAE}   & $0.0068$ & $0.0061$ & $0.0085$ & $0.0065$ & $0.0137$ \\
        \textit{RMSE}  & $0.0335$ & $0.2876(*)$ & $0.0311$ & $0.0290$ & $0.0346$ \\
        \textit{AEq95} & $0.0225$ & $0.0202$ & $0.0305$ & $0.0204$ & $0.0566$ \\
        \textit{AEq99} & $0.1164$ & $0.1176$ & $0.1359$ & $0.1195$ & $0.1667$ \\
        \textit{$R^2$} & $0.9954$ & $0.9966$ & $0.9960$ & $0.9965$ & $0.9951$ \\
        $\varepsilon_F$ & &  & $0.1158$  &$0.1142$ &$0.1457$\\
        \bottomrule
    \end{tabular}
    \caption{Pointwise error metrics on the test set for all three models. Results are grouped by source and model type. We believe \cite{hines2023graph} has a typo in (*), and possibly the RMSE of the GNN is 0.02876 instead of 0.2876.} 
    \label{tab:metricas_test_resultados}
\end{table}

\subsection{Generalisation performance across models: global metrics}
The global performance of the three models on the test set is summarised in \autoref{tab:metricas_test_resultados}, which also includes the reference values reported by \textcite{hines2023graph} for MLP and GNN. This comparison shows that the implementations developed here for both MLP and GNN yield metrics that are broadly consistent with those of the reference study --with GNN showing superior performance over MLP--. %probably arising from a few points with larger We attribute small discrepancies to factors such as randomness in weight initialization, stochasticity introduced during training, and differences in optimization trajectories.

\medskip \noindent 
The performance of all three surrogate models is arguably very good (e.g. reaching $R^2\ge 0.995$ in the three cases). This is a commendable result, reminding that the data split is performed only with respect to the operating conditions $(\mathscr{M}, \alpha)$: in this sense we are assessing the ability of the surrogate models to interpolate --and in some cases, also extrapolate-- over operating conditions, rather than over spatial locations on the airfoil's surface. Additionally, observe that the MSE metric on the test set is comparable (only slightly smaller) than the one on the validation set (\autoref{fig:training_curves_modelos}), overall certifying that the training and validation processes have been done correctly and that the data split in \autoref{fig:nlr7301_dataset} is adequate.

\medskip \noindent 
Now, when comparing the KAN model with the established MLP and GNN baselines, a small performance gap appears. The KAN exhibits slightly higher values across all error metrics, indicating a marginally inferior predictive accuracy. Recall, however, that the KAN model requires only about $10\%$ of the number of training epochs required by both MLP and GNN, hence KAN's comparatively inferior performance needs to be traded-off with its apparent improvement on training efficiency. We shall come back to this point later.
%Similarly, its $R^2$ score is noticeably reduced compared to the other models, suggesting that the KAN struggles to capture the nonlinear relationships present in the data with the same effectiveness. These results highlight that, under the current configuration and training setup, the KAN does not achieve the same level of generalization as the MLP and GNN, which remain the more reliable surrogates for this kind of problems.

\subsection{Generalisation performance across models: additional metrics}

\begin{figure}[H]
    \centering
    \begin{subfigure}{0.32\linewidth}
        \centering
        \includegraphics[width=\linewidth]{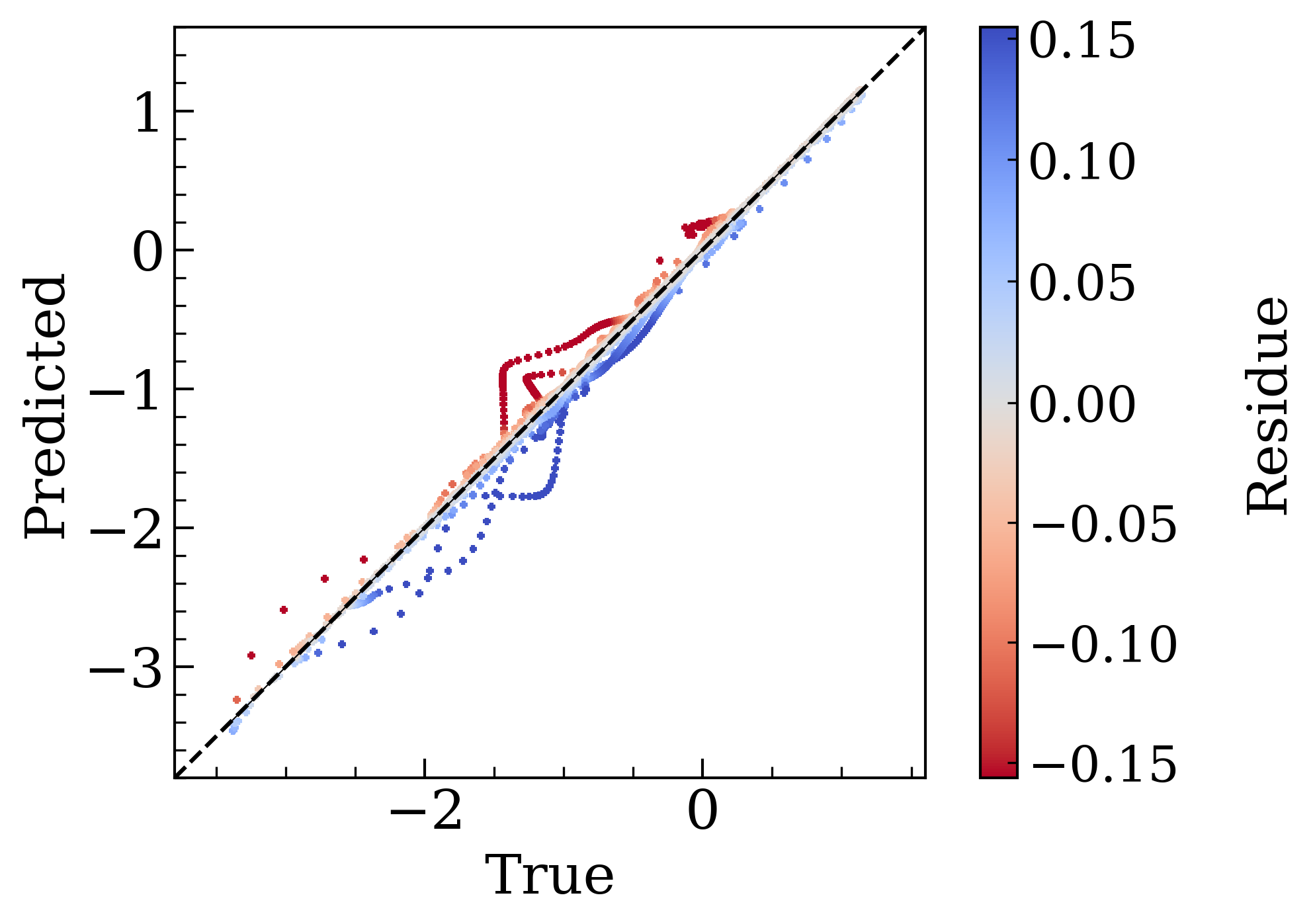}
        \caption{MLP.}
        \label{fig:true_pred_mlp}
    \end{subfigure} \hfill
    \begin{subfigure}{0.32\linewidth}
        \centering
        \includegraphics[width=\linewidth]{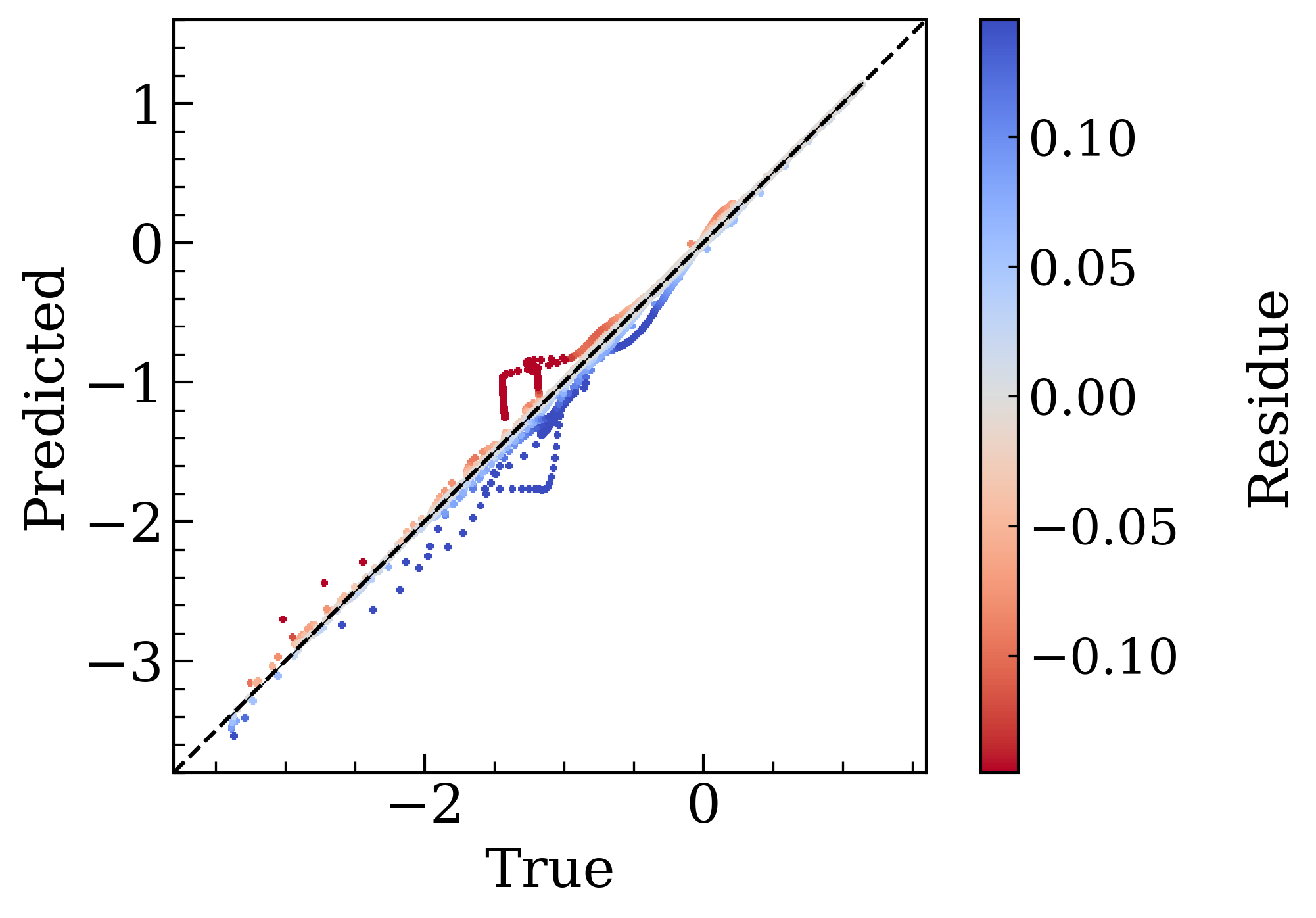}
        \caption{GNN.}
        \label{fig:true_pred_gnn}
    \end{subfigure} \hfill
    \begin{subfigure}{0.32\linewidth}
        \centering
        \includegraphics[width=\linewidth]{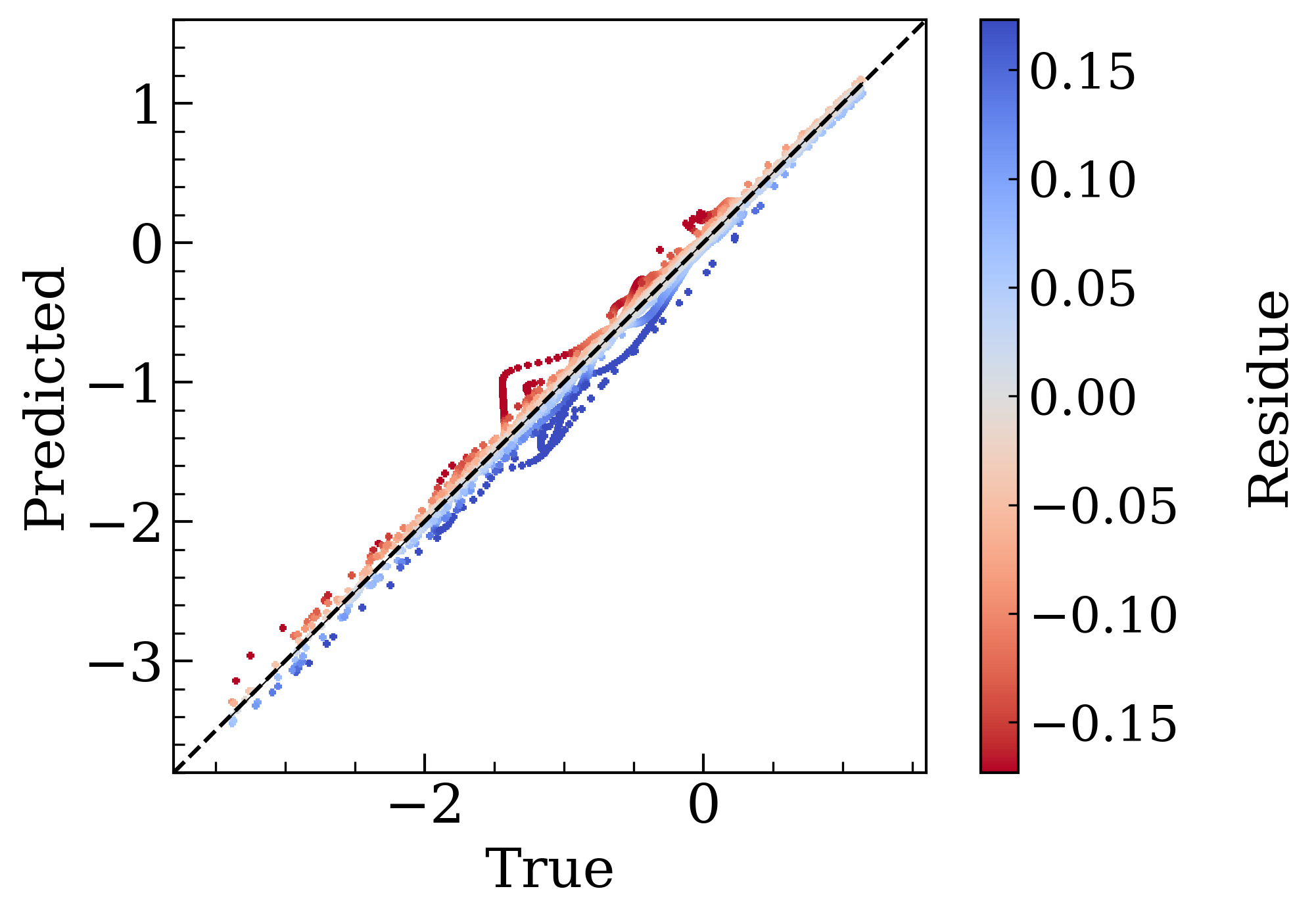}
        \caption{KAN.}
        \label{fig:true_pred_kan}
    \end{subfigure} 
    \caption{True versus predicted surface pressure coefficients for the three models. The color scale indicates the residual magnitude.}
    \label{fig:true_pred}
\end{figure}

In this section we provide a more detailed analysis of the generalisation performance of the three models. In the three panels of \autoref{fig:true_pred}, we plot scatter plots of the true pressure coefficient $C_p(x_i,z_k)$ vs its prediction $\hat{C}_p(x_i,z_k)$, for all location points $x_i$ and operating conditions $z_k=(\mathscr{M},\alpha)_k$ in the test set (the coefficients of determination $R^2$ are depicted in \autoref{tab:metricas_test_resultados}). We color-code each point $(C_p(x_i,z_k), \hat{C}_p(x_i,z_k))$ in terms of its residue value $C_p(x_i,z_k)- \hat{C}_p(x_i,z_k)$. For instance, points above the diagonal correspond to cases where $\hat{C}_p(x_i,z_k) > C_p(x_i,z_k)$. This kind of color-coded scatter-plots allows us to assess in what regions of the output space the model is more or less conservative, something important in safety-critical applications \cite{lacasa2025towards}. {Comparing across models (Panels a-c), the MLP and GNN models exhibit their largest deviations in the central region of the pressure coefficient distribution, particularly for values in the range of approximately $-2$ to $0$. These values are typically associated with the suction side (extrados) of the airfoil, where the flow acceleration leads to negative pressure coefficients. The presence of larger deviations in this range suggests that part of the underlying physics governing the flow in this region is not being adequately captured by the MLP and GNN models.
In turn, the KAN shows a slightly broader dispersion across the entire range of $C_p$, indicating difficulties in accurately capturing the pressure coefficient over all regions of the domain. This behavior is consistent with the global performance metrics, which are slightly inferior compared to those of the MLP and GNN models.}.

\begin{figure}[H]
    \centering
    \begin{tabular}{cc}
        \begin{subfigure}{0.47\textwidth}
            \centering
            \includegraphics[width=\textwidth]{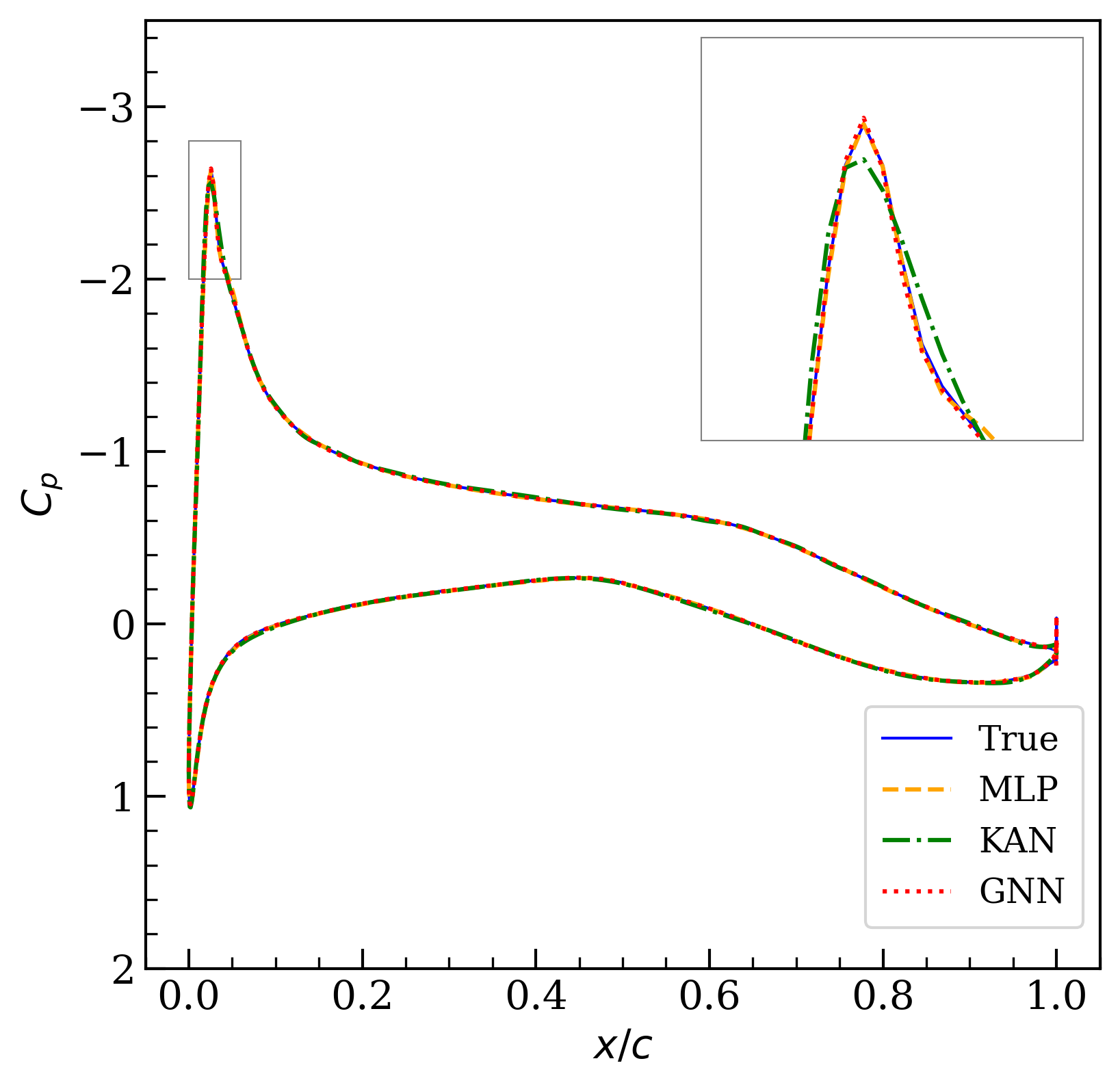}
            \caption{Subsonic interpolation case: \(\mathscr{M}=0.4441\), \(\alpha=3.123\) deg.}
            \label{fig:nlr-dlr_worst_1}
        \end{subfigure}
        &
        \begin{subfigure}{0.47\textwidth}
            \centering
            \includegraphics[width=\textwidth]{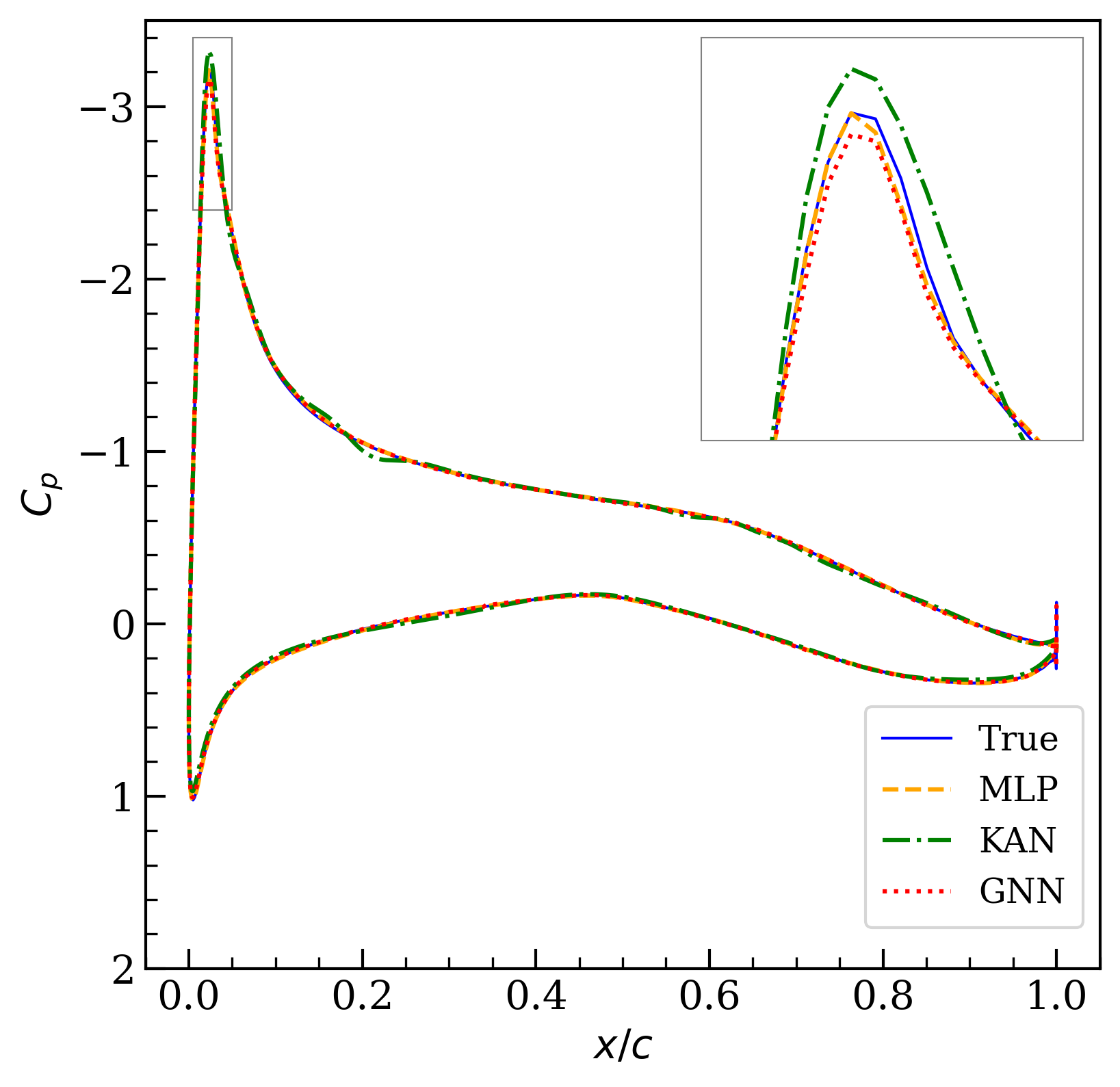}
            \caption{Subsonic extrapolation: \(\mathscr{M}=0.3176\), \(\alpha=4.901\) deg.}
            \label{fig:nlr-dlr_worst_2}
        \end{subfigure}
        \\
        [1em]
        \begin{subfigure}{0.47\textwidth}
            \centering
            \includegraphics[width=\textwidth]{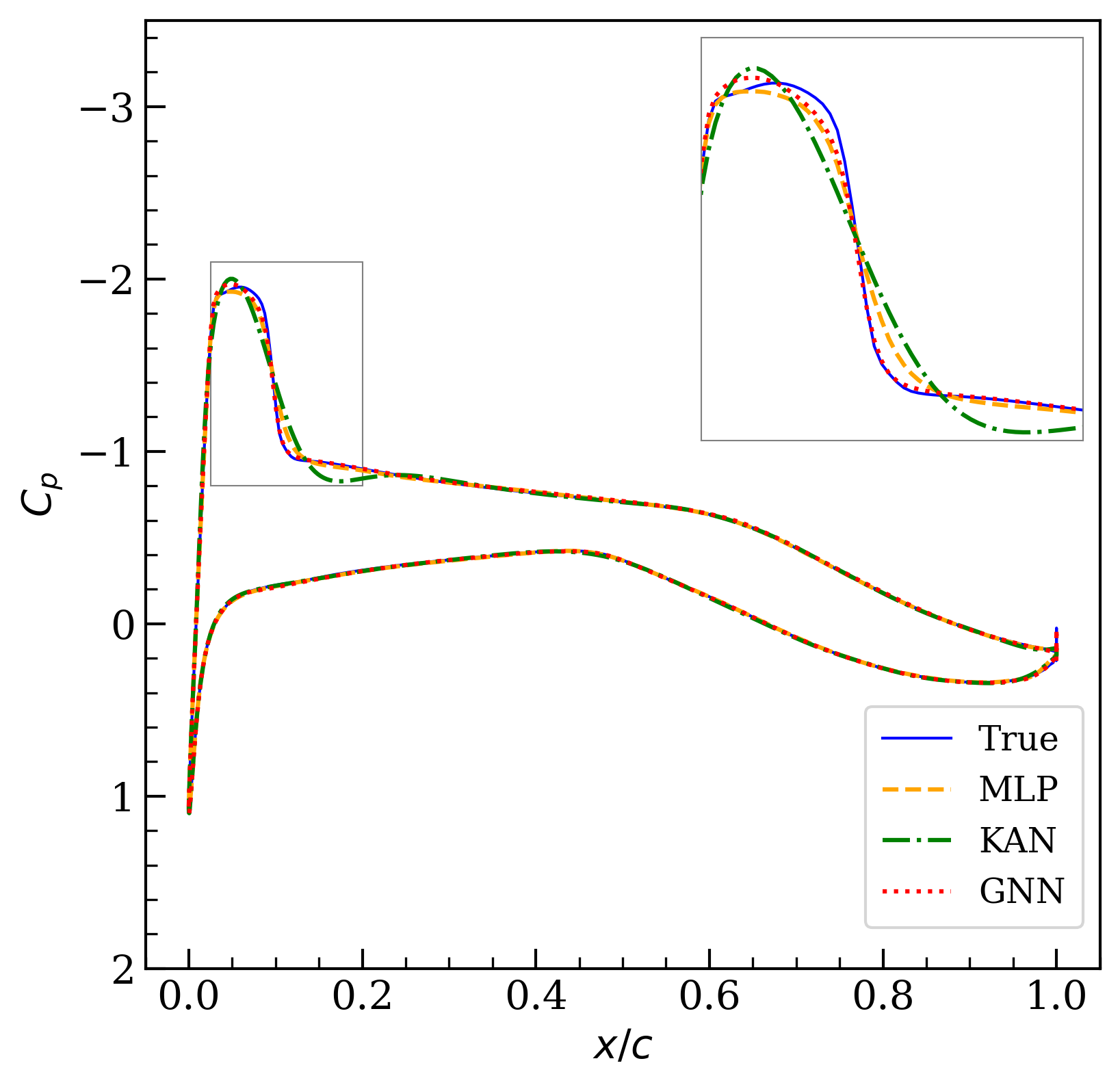}
            \caption{Transonic interpolation: \(\mathscr{M}=0.6304\), \(\alpha=1.642\) deg.}
            \label{fig:nlr-dlr_worst_3}
        \end{subfigure}
        &
        \begin{subfigure}{0.47\textwidth}
            \centering
            \includegraphics[width=\textwidth]{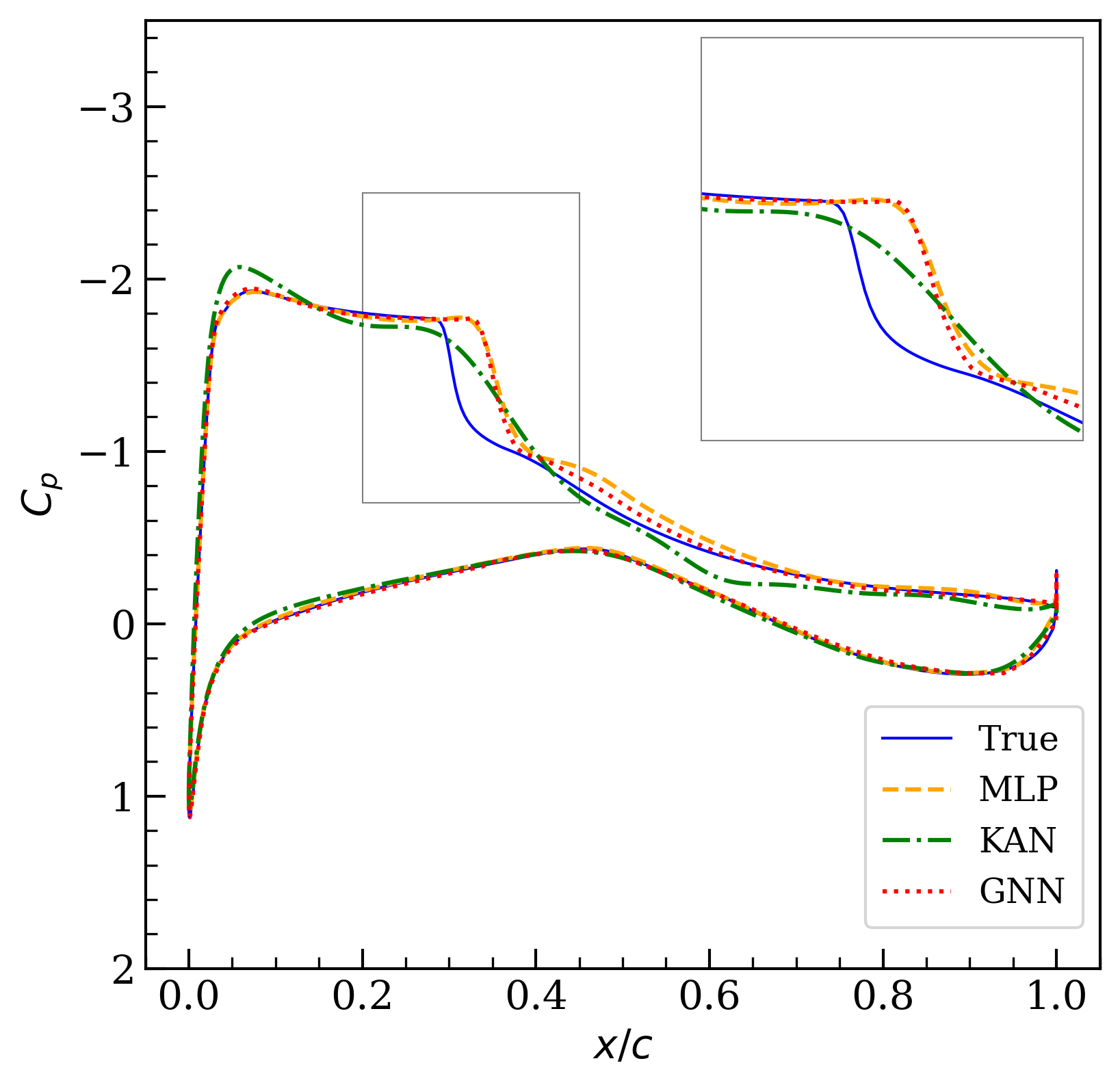}
            \caption{Transonic extrapolation: \(\mathscr{M}=0.6972\), \(\alpha=4.605\) deg.}
            \label{fig:nlr-dlr_worst_4}
        \end{subfigure}
    \end{tabular}
    \caption{Surface pressure distribution at four different operating conditions of the test set. Each panel additionally shows an inset with a zoom of a region of interest.}
    \label{fig:nlr-dlr_worst}
\end{figure}

To further understand where prediction errors are located within the airfoil, the pressure coefficient distributions for four operating conditions are presented in \autoref{fig:nlr-dlr_worst}, following the selection criteria of \textcite{hines2023graph}. The first two correspond to cases in the low subsonic regime, whereas the last two represent transonic conditions. These cases are chosen such that, for each regime, one case lies inside the convex hull of the training set (in the $\mathscr{M}-\alpha$ plane, hence interpolating operating conditions), and the other lies outside the convex hull, i.e. effectively extrapolating operating conditions.\\
In the subsonic interpolation case (\autoref{fig:nlr-dlr_worst_1}), all three models are able to reproduce the true pressure coefficient distribution with good accuracy, although the KAN slightly underestimates the suction peak near the leading edge. In contrast, for the subsonic extrapolation case (\autoref{fig:nlr-dlr_worst_2}), the KAN exhibits more pronounced deviations throughout the distribution compared to the MLP and GNN, indicating a slightly weaker extrapolation capability. For the transonic interpolation case in \autoref{fig:nlr-dlr_worst_3}, the three models successfully capture the shock that appears on the suction side of the airfoil, but the KAN overestimates the suction peak and exhibits a less accurate reconstruction of the pressure plateau downstream of the shock. Finally, the transonic extrapolation in \autoref{fig:nlr-dlr_worst_4} represents the most challenging scenario: none of the models accurately predicts the precise shape of the shock in the airfoil. The MLP and GNN tend to shift it downstream, but they represent the pressure jump with a more realistic gradient. In contrast, although the KAN predicts the geometric location on the suction side where the shock begins more accurately, it tends to excessively smooth the transition, leading to a more diffuse pressure gradient.

%In summary, \autoref{fig:nlr-dlr_worst} shows that all models achieve satisfactory results under subsonic conditions, although the KAN demonstrates a marginally poorer performance when extrapolating. \textcolor{orange}{FG: In the transonic regime, however, the accuracy of the KAN is slightly inferior to that of the other models, suggesting a limited ability to capture non-linear phenomena such as shock waves, especially in extrapolation scenarios. Seguimos pensando esto? Los nuevos resultados de la KAN la hacen bastante similar al MLP}

\begin{figure}[H]
    \centering
    \includegraphics[width=0.7\textwidth]{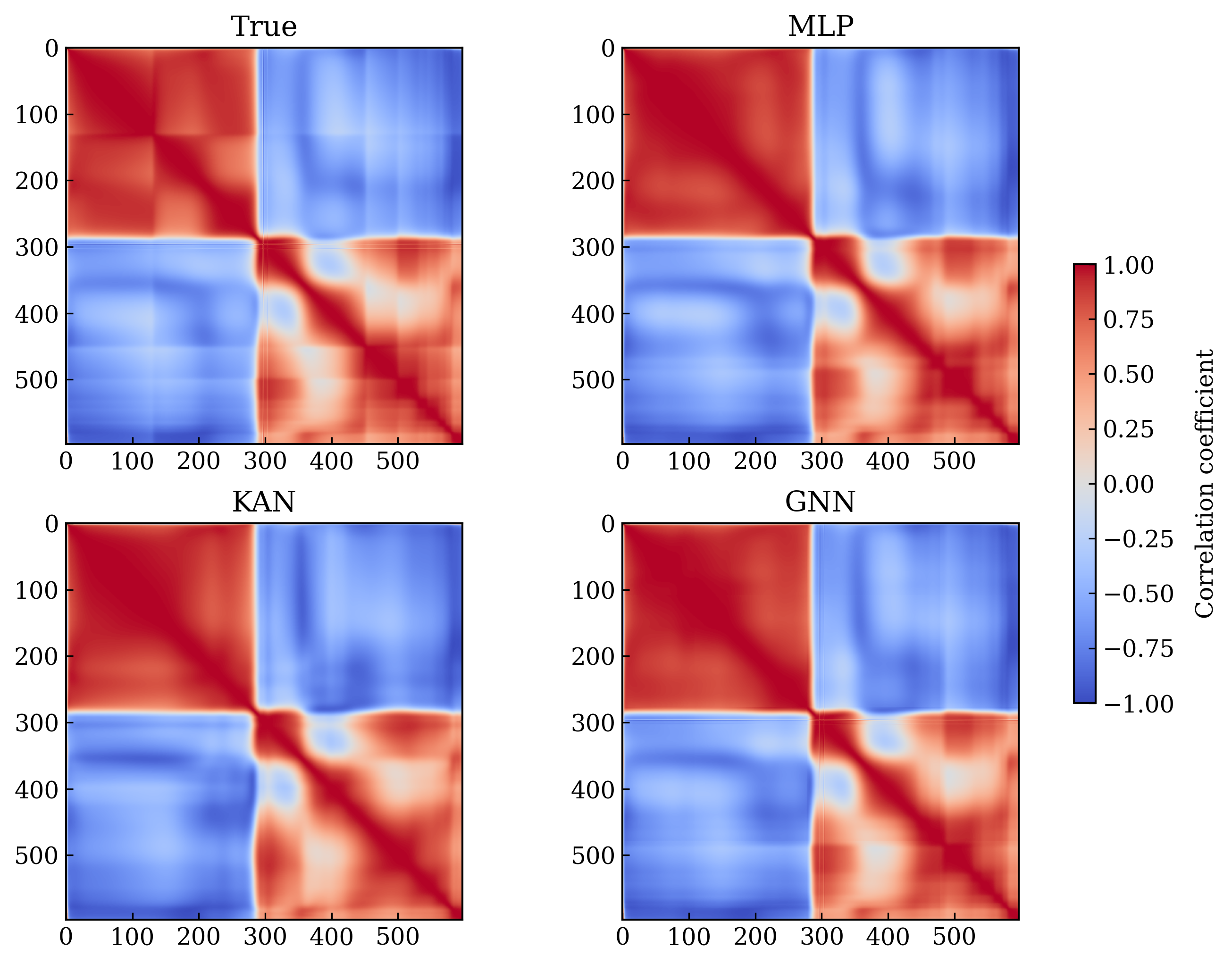}
    \caption{Correlation matrices for the test set: (top left) reference, (top right) MLP, (bottom left) KAN, and (bottom right) GNN.}
    \label{fig:correlation_matrices}
\end{figure}

To round-off the analysis, \autoref{fig:correlation_matrices} compares the test-set correlation matrices (Eq.~\ref{eq:corr}) for the ground truth and for the three surrogate models, whereas its  Frobenius relative error  $\varepsilon_F$ (Eq.~\ref{eq:frob}) is depicted in the last line of \autoref{tab:metricas_test_resultados}. All three models preserve the correlation structure, capturing the dominant structures and the transition zones. In comparison, the GNN model is the one that, overall (according to $\varepsilon_F$) better preserves the correlation structure among pressure coefficients over the airfoil. This is somewhat expected since, being a global predictor, this model explicitly uses such spatial information during training. In this sense, the performance of both the MLP and the KAN is commendable, given that these are trained as local regressors where no inductive bias related to the spatial continuity of $C_p$ is explicitly incorporated. 
%The KAN model, also operating as a local regressor, achieves a comparable level of accuracy, with only a slightly higher $\varepsilon_F$ and minor deviations in the reconstructed correlation patterns. Overall, both local approaches remain competitive, although they are marginally outperformed by the GNN in preserving the global correlation structure.}

\subsection{MLP vs KAN with comparable complexity}
In previous sections we have found that, while both MLP and KAN-based surrogate models achieve good performance on the learning task studied here, the performance of MLP is slightly yet systematically superior over the KAN for a variety of global and local metrics. At the same time, we also found that the optimal KAN model was much smaller than the optimal MLP one given in \cite{hines2023graph} (in terms of number of trainable parameters), and its training was accordingly significantly faster. Thus, it remains difficult to conclude the supremacy of one against the other, since a direct comparison of the two optimised models is not really fair as their model complexities are very different (in particular, the number of hidden layers of the optimal MLP is much larger than for the KAN model).\

\begin{figure}[htb!]
    \centering
    \includegraphics[width=0.5\linewidth]{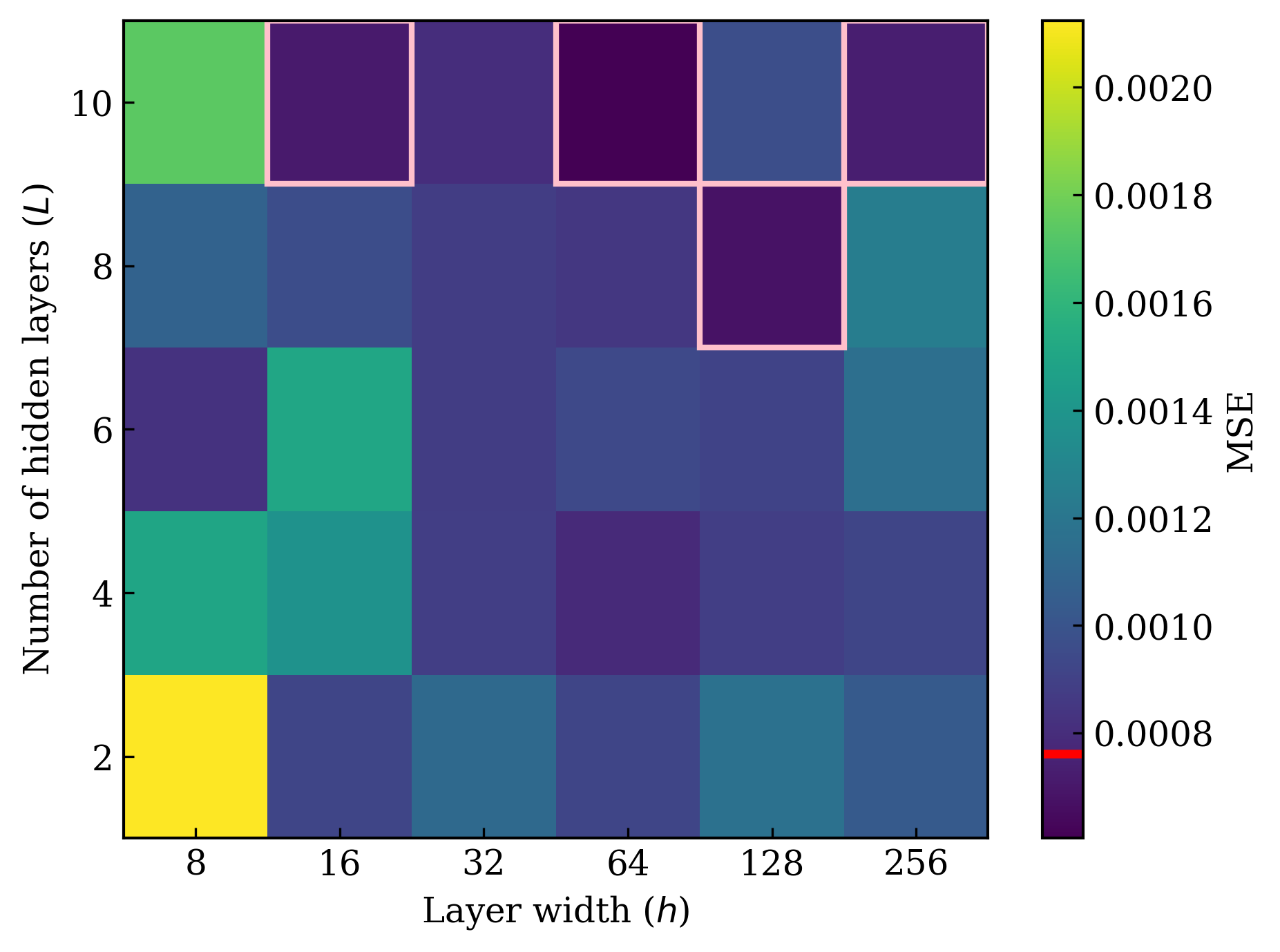}
    \includegraphics[width=0.4\linewidth]{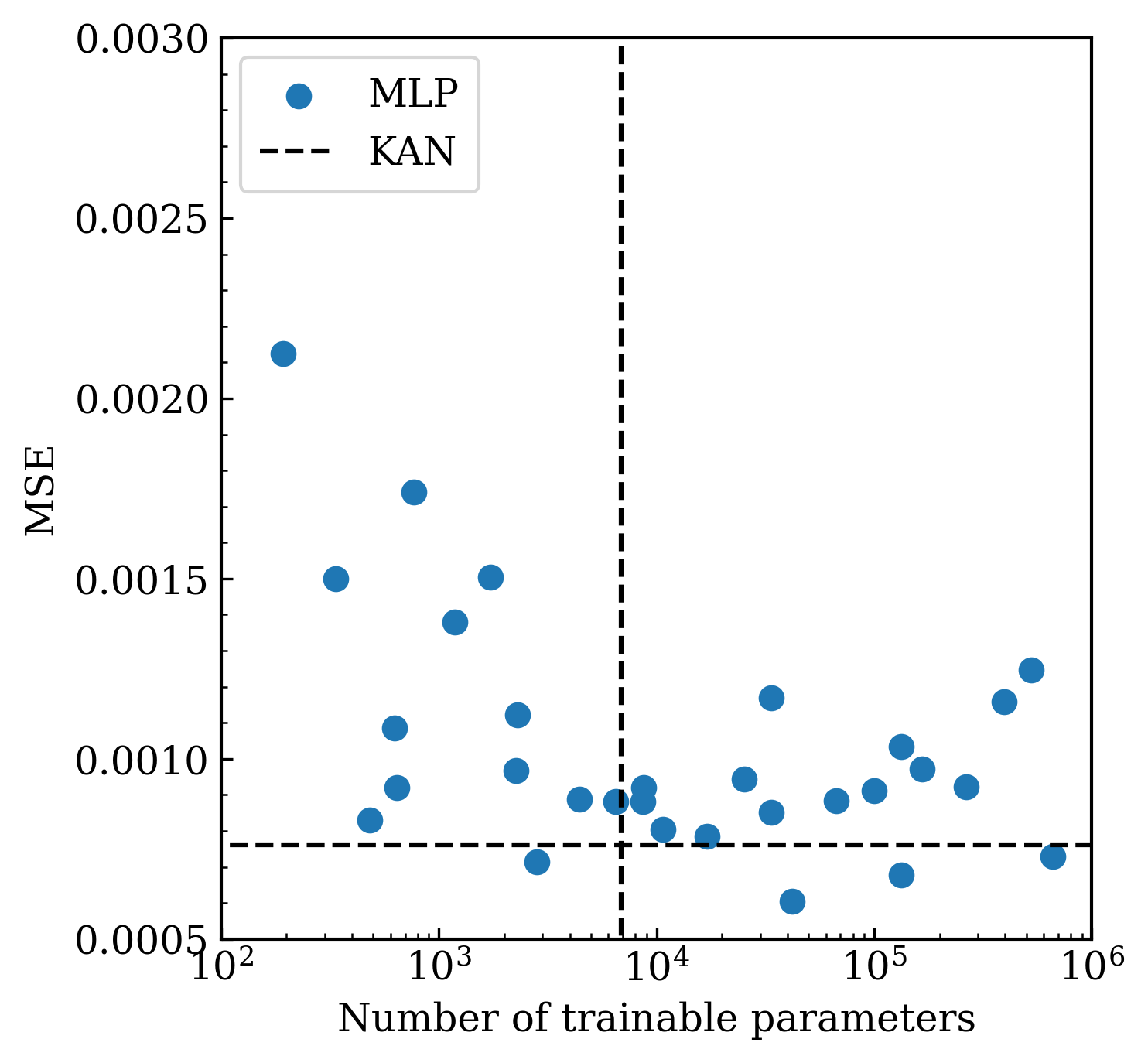}
    \caption{Comparison of the KAN model against a MLP of comparable complexity. (Left panel): Heatmap of the validation set MSE of a MLP model as a function of its model's complexity (number of hidden layers and layer width). The cases exceeding the performance level of the KAN are highlighted in pink, with this reference level indicated in the color bar by a red line. (Right panel): Validation set MSE of a MLP model, as a function of its model complexity (total number of trainable parameters). For reference, the complexity of the optimal KAN model and its validation MSE is highlighted (dashed lines).} %\textcolor{magenta}{En la de la derecha, sacar el training time (número de epochs o similar) para uno comparable en error a la KAN para comparar coste de entrenamiento.}\textcolor{blue}{LL intuitivamente yo diría que todos los modelos MLP que tengan menos parámetros entrenables tardarán menos (y casi todos además tienen menor error? Pero es cierto que no necesariamente es solo el número de parámetros lo que determina el número de epochs necesarias hasta convergencia... Fermín has usado un número de epochs fijo en el MLP? Si es así es un poco rollo, tendrías que pintar learning curves y mirar caso por caso a ver en realidad cuándo la learning curve ha convergido a un valor similar al que pintas en esa gráfica...} \textcolor{magenta}{si, la intución mía tb es esa. Y por reducir la carga de trabajo, comentaba la posibilidad de sacar el training time solo para un caso.} \textcolor{orange}{FG: UPDATED. Respecto a los tiempos de entrenamiento y el numero de epochs, al final creo que quedamos en no mencionar nada. No obstante, recuerdo que todos los MLP de este estudio se entrenaron 500 epocas y se optimizó con optuna (50 trials para cada modelo) el learning rate, learning rate decay factor, y el batch size. Y revisando los entrenamientos, todos los modelos han convergido correctamente.}}
    \label{fig:MLP_SMALL}
\end{figure}

\medskip \noindent
To be able to make a fair comparison, here we depart from the optimal MLP configuration given in \cite{hines2023graph} and compare the performance of MLP vs KAN for smaller MLP models (in terms of number of hidden layers and/or number of neurons per layer), so that the overall complexity of both models (in terms of total number of trainable parameters) is made comparable. Results are shown in \autoref{fig:MLP_SMALL}, and certify that when we reduce the complexity of the MLP model --and thus reducing its training requirements--, its performance (measured in the validation set) is similar to the KAN model although typically a bit worse. Observe that the best MLP are the ones that preserve a large layer depth (11 hidden layers). In other words, it seems that the expressivity of the MLP comes from its layer depth, something which is in contrast to the KAN model, that based on the hyperparameter optimisation operates with only two hidden layers. This analysis indicates that while MLP might seem more suitable than KAN if we only consider performance, if we are restricted to low-complexity models (i.e. due to restrictions in computational resources), a KAN model might be preferred, provided that we carefully optimise its hyperparameters.
%.\textcolor{red}{LL ES CIERTO QUE SOLO MIRANDO EL NUMERO DE PARAMETROS NO PODEMOS SABER CONCRETAMENTE EL NUMERO DE EPOCHS NECESARIA PARA CONVERGER, PARA QUE ESTUVIERA ESTO TOTALMENTE REDONDO HABRÍA QUE PINTAR ESE NUMERO DE EPOCHS NECESARIO PARA CADA MODELO MLP... } \textcolor{orange}{Aquí se podría comentar algo de que, obervando la \autoref{fig:MLP_SMALL}, y pese a tener un numero total de parametros entrenables inferior, los MLP que mejoran los resultados de las KAN son los que tienen arquitecturas muy profundas (9 u 11 capas ocultas). Me parece llamativo este dato, teniendo en cuenta que para la KAN la arquitectura óptima es de 2 capas ocultas, todo lo contrario.}

\section{Conclusions}

\label{sec:conclusions}
All three deep learning surrogate models constructed in this work --a multilayer perceptron, a graph neural network and a Kolmogorov-Arnold network-- showed very good performance in the prediction of pressure coefficients alongside the boundary of an airfoil, and were able to accurately interpolate across different Mach numbers and angles of attack. Among the three models, typically the GNN showed the best results, followed by the MLP. The KAN model was marginally inferior to the MLP, hence providing relevant insights on the debate around the KAN supremacy over MLP \cite{yu2024kan}.\\
Incidentally, the training of KAN model proved to be significantly more challenging than that of the other architectures considered in this study. In particular, the optimization process exhibited notable instabilities during training, frequently leading to extremely large gradients. This behavior complicated the learning process and made the optimization of hyperparameters substantially more challenging. As a consequence, achieving stable convergence requires careful tuning and increases the computational effort associated with training these models.\\
%An additional observation concerns the nature of the optimal hyperparameters obtained during model calibration. In many cases, the best-performing KAN configurations tended to favor low-order Chebyshev polynomials, typically of order one or two. From a practical standpoint, such low-order functions significantly reduce the expressive differences between KANs and more traditional neural architectures. In fact, when restricted to such low-order representations, the resulting model behaves similarly to a classical multilayer perceptron, effectively diminishing the supposed theoretical advantages associated with the functional representation employed by KANs.\\
Regarding predictive performance, as mentioned previously all three models showed notably low generalisation (test) error, but KANs exhibited slightly inferior results. 
Regions with strong pressure gradients along the airfoil surface were more difficult to accurately predict for all cases.\\
%In particular, that model showed greater difficulty accurately reproducing regions of the solution characterised by strong gradients, such as those typically associated with rapid pressure variations along the airfoil surface. 
%This behavior is not unique to KANs, as similar challenges were observed for the other models analyzed, including MLPs and graph neural networks (GNNs). However, among all the architectures considered, KANs demonstrated the weakest performance precisely in these high-gradient regions, which are often the most critical for aerodynamic analysis.\\
One positive property of KANs over both MLPs and GNNs is that its optimal model complexity (in terms of number of training parameters) is orders of magnitude smaller than the MLP and GNN ones, and thus the number of epochs needed to achieve good performance is systematically smaller in the KAN case --what results in faster training-- when comparing optimal model configurations. When comparing MLP and KAN models of similar number of trainable parameters, the KAN architecture showed to be competitive and typically at least as good as the MLP.
%Furthermore, it was observed that the level of predictive accuracy achieved by the KAN models could generally be matched by a standard MLP architecture with a smaller number of trainable parameters. This suggests that, for the class of problems considered here, the additional architectural complexity introduced by KANs does not translate into a corresponding improvement in model efficiency or predictive capability.\\

Taken together, these results provide a nuanced view on the MLP vs KAN debate in the context of building fluid-dynamics-based surrogate models, with no clear supremacy. Classical deep learning architectures such as MLP or GNN show marginally superior or marginally inferior performance as compared to KAN when considering different aspects such as training stability and efficiency, overall performance, or performance restricted to low-complexity models. Further work should clarify whether these insights hold beyond aerodynamic cases within fluid-dynamics-based surrogate modeling.
%indicate that Kolmogorov–Arnold Networks are not the most suitable surrogate modeling approach for the type of aerodynamic regression problems investigated in this study. Despite their appealing theoretical foundations, their practical advantages appear limited in this context, particularly when compared with simpler and more stable architectures such as multilayer perceptrons.

\medskip \noindent {\bf Code --} All stages of the pipeline followed throughout this work, including training, hyperparameter optimization, and evaluation of the three surrogate models has been executed on the FLEXO cluster at the Universidad Politécnica de Madrid (UPM), which includes a computing node equipped with 4 AMD Instinct MI210 GPUs and 80 Intel(R) Xeon(R) Silver 4416+ CPU cores. A single GPU and 10 CPU cores were used for each model, allowing parallel data loading and preprocessing. Code details are available {at \url{https://github.com/ArnauMiro/pyLowOrder.git} \cite{pyLOM}, where the implementation of the model architectures is provided.}

\medskip \noindent
{\bf Acknowledgments --- }We thank Derrick Hines and Philipp Bekemeyer (DLR) for sharing their database \cite{hines2023graph} and for insightful discussions. The authors acknowledge funding from project TIFON (PLEC2023-010251) funded by
MCIN/AEI/10.13039/501100011033, Spain. 
The authors acknowledge funding from the European Union (project HERFUSE) under GA No 101140567. Views and opinions expressed are however those of the author(s) only and do not necessarily reflect those of the European Union or Clean Aviation Joint Undertaking. Neither the European Union nor Clean Aviation JU can be held responsible for them.
LL acknowledges partial support from project CSxAI (PID2024-157526NB-I00) funded by MICIU/AEI/10.13039/501100011033/FEDER, UE, project Maria de Maeztu (CEX2021-001164-M) funded by the MICIU/AEI/10.13039/501100011033, and from the European Commission Chips Joint Undertaking project No. 101194363 (NEHIL).
GR acknowledges partial financial support received by the Grant DeepCFD (Project No. PID2022-137899OB-I00) funded by MICIU/AEI/10.13039/501100011033 and by ERDF, EU. 
Finally, all authors gratefully acknowledge the Universidad Politécnica de Madrid for providing computing resources on Magerit Supercomputer.

\printbibliography
%\bibliography{bibliography}

%\include{apendice}

\end{document}